\documentclass[11pt]{article}
\usepackage[margin=1in]{geometry}
\usepackage{amsmath,amssymb,amsthm}
\usepackage{graphicx}
\usepackage{booktabs}
\usepackage{subcaption}
\usepackage{hyperref}
\usepackage{xcolor}

\newtheorem{observation}{Observation}

\title{\textbf{Statistical Inference and Quality Measures of KV Cache Quantisations Inspired by TurboQuant}}
\author{Paolo D'Alberto\thanks{These notes were prepared in collaboration with Claude (Anthropic).}\\
\small\texttt{paolodalberto@amd.com}}
\date{}

\begin{document}
\maketitle

\begin{abstract}
We analyse three KV cache quantization schemes under a fair bit budget:
\textbf{KV} (scalar MSE baseline), \textbf{KQV} (WHT + MSE on $K$;
WHT + MSE + QJL on $V$), and \textbf{QKQV} (WHT + MSE + QJL on both).
Starting from the Beta distribution on the hypersphere, we trace how
QJL on $K$ inflates inner product variance by $\pi/2$, which softmax
amplifies nonlinearly via Jensen's inequality.

Three empirical findings emerge.
(1)~At $n=4$ (the practically dominant budget), KQV wins on every
measure --- KL divergence, geometric $K$ error, and 6D distance ---
across all distributions and ranks tested.
(2)~The K--V asymmetry is unconditional: QKQV is consistently worse
than KQV in KL divergence at every budget and distribution.
(3)~A budget-dependent crossover exists: QKQV achieves better geometric
$K$ reconstruction at $n \in \{2,3,5\}$, KQV at $n \in \{4,6\}$,
invariant to rank and tail weight --- an open rate-distortion problem.

$\mathrm{KL}(p_{\mathrm{ref}} \| p_{\mathrm{quant}})$, K-only by
construction, bridges K direction error to routing corruption and output
collapse. At $n=4$, KQV achieves $2.25\times$ lower KL than QKQV
(MW $r = -0.983$, $p < 10^{-23}$), explaining the crossover.
We introduce a 6D error framework providing distributional discrimination
and geometric resolution beyond scalar metrics.
\end{abstract}

\section{The Beta Distribution on the Hypersphere}
\label{lab:beta}

Let $\mathbf{x}$ be uniformly distributed on $\mathcal{S}^{d-1}$.
Each coordinate $x_j$ has density
\begin{equation}
  f_X(t) = \frac{\Gamma(d/2)}{\sqrt{\pi}\,\Gamma\!\left(\tfrac{d-1}{2}\right)}
            \left(1 - t^2\right)^{\frac{d-3}{2}},
  \quad t \in [-1,1],
\label{eq:beta}
\end{equation}
which is Beta$\!\left(\tfrac{d-1}{2},\tfrac{d-1}{2}\right)$ on $[-1,1]$.
As $d \to \infty$, this converges to $\mathcal{N}(0, 1/d)$.

A geometric proof follows.  Fix $x_1 = t$. The remaining coordinates
satisfy $x_2^2 + \cdots + x_d^2 = 1 - t^2$, forming a $(d-2)$-sphere
of radius $\sqrt{1-t^2}$ in $\mathbb{R}^{d-1}$.  The surface area
element at value $t$ is
\begin{equation}
  \begin{split}
    |dS\big|_t &\equiv |\mathcal{S}^{d-2}| \cdot (1-t^2)^{\frac{d-2}{2}}
    \cdot \frac{dt}{\sqrt{1-t^2}} \\
    &= |\mathcal{S}^{d-2}| \cdot (1-t^2)^{\frac{d-3}{2}}\,dt,
    \end{split}
\end{equation}
where the factor $1/\sqrt{1-t^2}$ is the arc-length correction
(the sphere curves away from the vertical axis).
Normalising by $|\mathcal{S}^{d-1}|$ and substituting
$|\mathcal{S}^{k}| = 2\pi^{(k+1)/2}/\Gamma\!\left(\tfrac{k+1}{2}\right)$
yield Equation~\eqref{eq:beta}.

Both surface areas are expressed via Gamma functions:
$|\mathcal{S}^{d-1}| = 2\pi^{d/2}/\Gamma(d/2)$ and
$|\mathcal{S}^{d-2}| =
2\pi^{(d-1)/2}/\Gamma\!\left(\tfrac{d-1}{2}\right)$.  Their ratio is
the normalisation constant in Equation~\eqref{eq:beta}.  The Gamma
function arises naturally from the Gaussian radial integral
$\int_0^\infty r^{d-1}e^{-r^2}dr = \Gamma(d/2)/2$, which is the
universal {\em size unit} for $d$-dimensional sphere geometry.  Each
time a coordinate is fixed and the sphere projected, the dimension
drops by one and the Gamma argument shifts by $1/2$: from $d/2$ to
$(d-1)/2$. The density~\eqref{eq:beta} is precisely that dimensional
accounting.

\begin{figure*}[htbp]
  \centering
  \includegraphics[width=\textwidth]{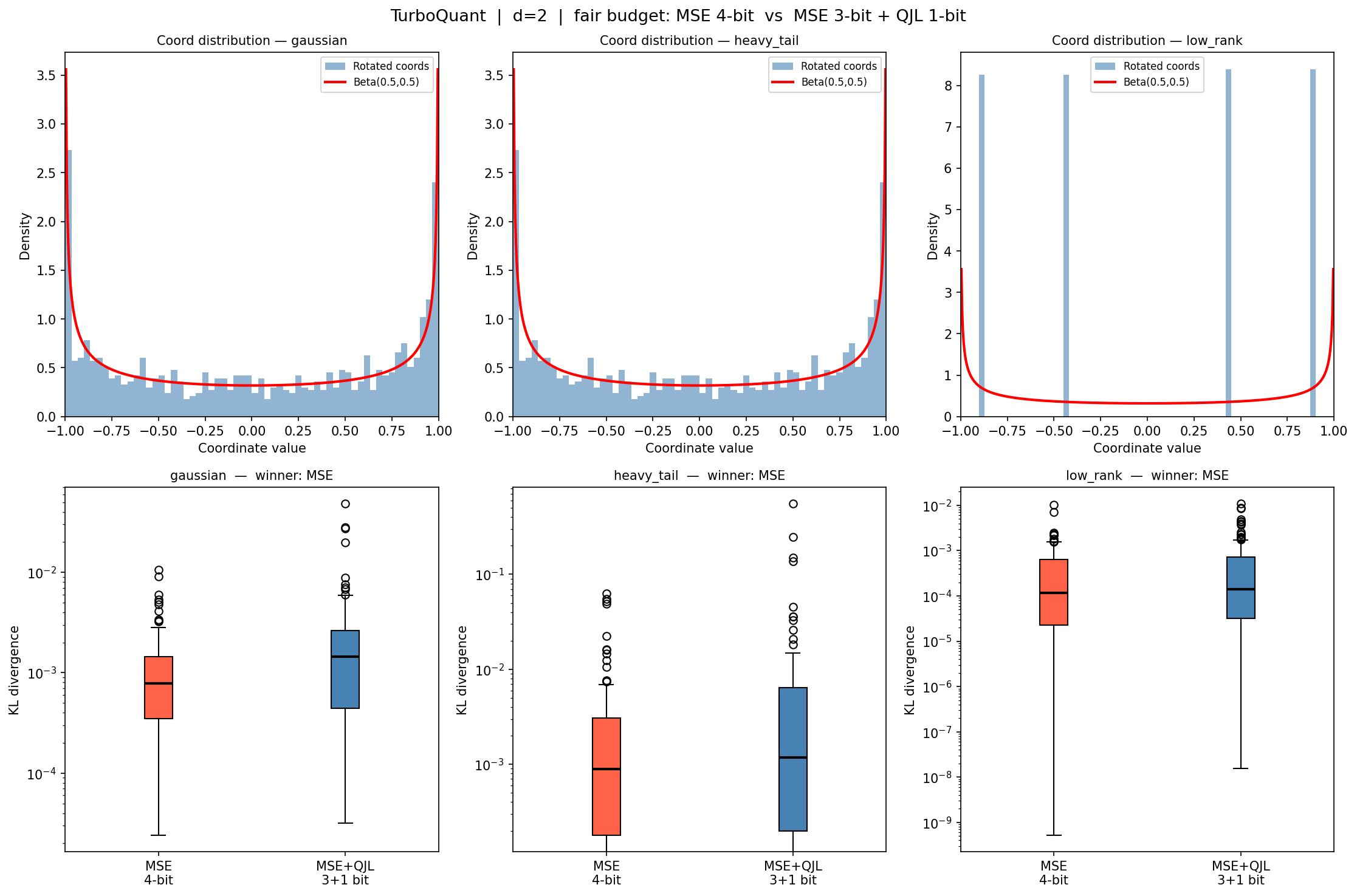}
  \caption{%
    $d=2$, Beta$(0.5,0.5)$ — arcsine (U-shaped).
    \textit{Top}: rotated coordinate histograms (blue) vs.\ Beta density
    (red) for \texttt{gaussian}, \texttt{heavy\_tail}, \texttt{low\_rank}.
    \textit{Bottom}: KL divergence boxplots, MSE 4-bit vs.\
    MSE 3-bit\,$+$\,QJL 1-bit (equal budget). MSE wins across all
    distributions. Circles denote outliers beyond $1.5\times\mathrm{IQR}$
    (${\approx}2.7\sigma$ for a Gaussian).
  }
  \label{fig:d2}
\end{figure*}

\begin{figure*}[htbp]
  \centering
  \includegraphics[width=\textwidth]{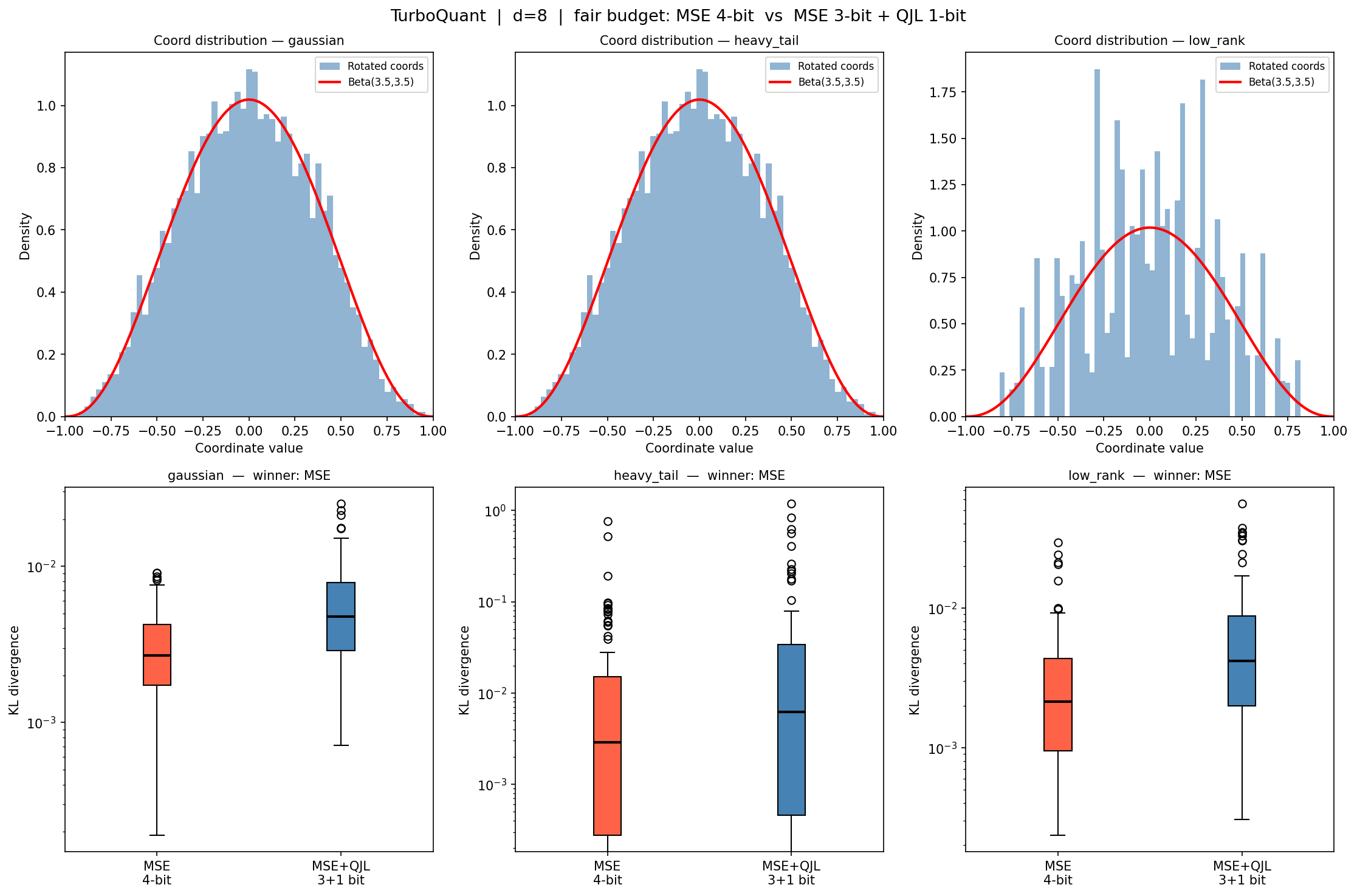}
  \caption{%
    $d=8$, Beta$(3.5,3.5)$ — mild bell.
    \textit{Top}: coordinate histograms begin to concentrate near zero;
    \texttt{low\_rank} (rank\,$=$\,1) shows a nearly flat histogram,
    deviating from Beta — the first sign of joint-structure failure.
    \textit{Bottom}: MSE wins across all distributions under equal budget.
    Circles denote outliers beyond $1.5\times\mathrm{IQR}$
    (${\approx}2.7\sigma$ for a Gaussian).
  }
  \label{fig:d8}
\end{figure*}

For large $d$ and small $t$, $(1-t^2)^{(d-3)/2} \approx e^{-dt^2/2}$,
a Gaussian with variance $1/d$.  This convergence is empirically
visible: the coordinate histogram is U-shaped (arcsine) at $d=2$, a
mild arch at $d=4$, a bell at $d=8$, and indistinguishable from
Gaussian at $d=1024$.  Figures~\ref{fig:d2}, \ref{fig:d8}, \ref{fig:d128}, and \ref{fig:d1024} show this
progression across four representative dimensions. Notice that the
Beta distribution is defined in the interval $[-1,1]$ and the Gaussian
for any finite $d$ is not and it can only be approximated to.

\section{The TurboQuant Pipeline}
In transformer attention, the key and value tensors computed at each
decoding step are stored and reused for all future steps; this memory
store is the \emph{KV cache}. Its size grows linearly with context
length, making it the dominant memory bottleneck at inference time and
the primary target for compression.

TurboQuant~\cite{zandieh2025turboquant} compresses KV cache vectors through a two-stage pipeline.
The first stage applies a structured random rotation followed by a
Beta-optimal scalar quantiser; the second stage corrects the inner
product bias in the residual via a 1-bit QJL sketch. Together the two
stages target orthogonal problems — variance and bias — as we describe
in turn.

\subsection{Stage 1: Rotation and Beta-Optimal Quantisation}
A random Hadamard--Rademacher transform $\mathbf{x}_{\mathrm{rot}} =
\tfrac{1}{\sqrt{d}}\,H\,\mathrm{diag}(\mathbf{s})\,\hat{\mathbf{v}}$
(where $\hat{\mathbf{v}} = \mathbf{v}/\|\mathbf{v}\|$ and $\mathbf{s}
\in \{-1,+1\}^d$ is a random sign vector) is applied before
quantisation. As shown in Section~\ref{lab:beta}, each coordinate
of $\mathbf{x}_{\mathrm{rot}}$ follows the Beta distribution, enabling
a scalar Lloyd-Max quantiser~\cite{max1960quantizing,lloyd1982least} to be designed once for this known
density and applied coordinate-wise. The original norm
$\|\mathbf{v}\|$ is stored separately. This minimises the mean squared
error (MSE) of the quantised representation.

\textbf{Limitation.} The random rotation is a single fixed mapping,
not a per-vector fresh rotation. It moves data off the coordinate axes
and provides expected uniformity over the randomness of $\mathbf{s}$,
but it cannot make a non-isotropic distribution isotropic. In other
words, the Hadamard transform with proper normalisation is an
orthonormal transformation: applied to unit vectors on the sphere, it
performs a rotation. The randomisation, if independent of the process
generating the vectors, is introduced to disrupt pathological
axis-aligned patterns (e.g., the vector $(1,0,\dots,0)$).

For a Gaussian input the Beta guarantee holds; for low-rank or
subspace-concentrated inputs the marginal coordinates may appear Beta
while the joint structure remains degenerate. We reproduce such
pathological but realistic embedding distributions and show where
Equation~\eqref{eq:beta} fails in practice.

\subsection{Stage 2: QJL Residual Correction}
Let $\mathbf{r} = \mathbf{x}_{\mathrm{rot}} - \hat{\mathbf{x}}_{\beta}$
be the Stage~1 residual. The 1-bit QJL sketch~\cite{zandieh2024qjl}
$\mathbf{q} = \mathrm{sign}\!\left(H(\mathbf{s}' \odot \mathbf{r})\right)$
with a fresh sign vector $\mathbf{s}'$ yields the unbiased estimator
\begin{equation}
  \hat{\mathbf{r}} = \frac{\sqrt{\pi/2}\,\|\mathbf{r}\|}{d}\,
                     \mathbf{s}' \odot H\mathbf{q},
  \qquad
  \mathbb{E}[\langle \mathbf{q}, \hat{\mathbf{r}}\rangle]
  = \langle \mathbf{q}, \mathbf{r}\rangle
  \quad \forall\,\mathbf{q}.
\end{equation}
The constant $\sqrt{\pi/2}$ corrects for the sign-quantisation bias:
for $z \sim \mathcal{N}(0,\sigma^2)$,
$\mathbb{E}[|z|] = \sigma\sqrt{2/\pi}$,
so $\mathrm{sign}(z)$ underestimates magnitude by $\sqrt{2/\pi}$.

The unbiasedness of $\hat{\mathbf{r}}$ is \emph{unconditional}: it holds
for any residual vector $\mathbf{r}$, regardless of whether Stage~1 is
optimal. What Stage~1 optimality affects is $\|\mathbf{r}\|$, and hence
the variance of the QJL estimate, which is proportional to
$\|\mathbf{r}\|^2/d$.

Note that Stage~2 applies the same Hadamard structure as Stage~1,
but with a fresh randomisation $\mathbf{s}'$ drawn independently of
$\mathbf{s}$, ensuring the sketch is uncorrelated with the
quantisation error.

\subsection{Three Schemes Under Study}
\label{sec:schemes}

All experiments compare three schemes at equal effective bit budget
$n$ per cache vector.  We adopt the following shorthand throughout:

\begin{table}[h]
\centering
\small
\caption{The three quantisation schemes compared throughout this paper.}
\label{tab:schemes}
\begin{tabular}{llll}
\toprule
Name & $K$ cache & $V$ cache & WHT \\
\midrule
\textbf{KV}   & MSE, $n$ bits        & MSE, $n$ bits                      & No  \\
\textbf{KQV}  & MSE, $n$ bits        & MSE $(n{-}1)$ bits + 1-bit QJL     & Yes \\
\textbf{QKQV} & MSE $(n{-}1)$ bits + 1-bit QJL & MSE $(n{-}1)$ bits + 1-bit QJL & Yes \\
\bottomrule
\end{tabular}
\end{table}

\textbf{KV} is the scalar baseline: no rotation, no sketch.
\textbf{KQV} applies WHT to both caches, then QJL as a residual
corrector on $V$ only — this is the scheme described in the original
TurboQuant paper as applied to inner products.
\textbf{QKQV} extends the QJL sketch symmetrically to $K$ as well.
All three store the exact vector norm as a float32; all remaining bits
are used for quantisation.  The Shannon reference
$2^{2n}$ is the same for KQV and QKQV at every budget $n$, making
the comparison fair.

The pairwise contrasts each isolate one mechanism:
KQV vs KV isolates WHT + QJL-on-$V$;
QKQV vs KQV isolates the effect of adding QJL to $K$;
QKQV vs KV isolates the full WHT + QJL package.

\section{Interaction with Softmax Attention}

In $\mathrm{softmax}\!\left(QK^T/\sqrt{d}\right)V$, the tensors $K$ and
$V$ play fundamentally different roles: $K$ enters only through inner
products $\langle \mathbf{q}, \mathbf{k}_i\rangle$, while $V$ enters as
a linear weighted sum. This asymmetry justifies optimising $K$
quantisation for inner product distortion rather than MSE.

\subsection{Jensen Bias in the Exponential}
Even with zero-mean inner product error
$\epsilon_i = \langle \mathbf{q}, \hat{\mathbf{k}}_i\rangle
             - \langle \mathbf{q}, \mathbf{k}_i\rangle$,
the convexity of $e^x$ introduces a bias:
\begin{equation}
  \mathbb{E}\!\left[e^{\langle \mathbf{q},\mathbf{k}_i\rangle + \epsilon_i}\right]
  = e^{\langle \mathbf{q},\mathbf{k}_i\rangle} \cdot e^{\sigma_i^2/2},
\end{equation}
for $\epsilon_i \sim \mathcal{N}(0,\sigma_i^2)$.
If $\sigma_i^2$ is \emph{uniform} across keys the factor $e^{\sigma^2/2}$
cancels in the softmax ratio.
The dangerous regime is when $\sigma_i^2$ is \emph{correlated with the
attention score}: high-attention keys with large quantisation variance
produce asymmetric distortion that cannot cancel.

\subsection{Role of QJL}
Stage~2 ensures $\mathbb{E}[\epsilon_i] = 0$ for every key,
preventing systematic drift in attention logits.
Stage~1 minimises $\sigma_i^2$, reducing the Jensen inflation.
Together they target the non-uniformity that would corrupt dominant
attention weights most.

\subsection{Per-Instance Bias and the Softmax Cancellation Argument}
After WHT and Lloyd-Max quantisation, the inner product error for a
specific key-query pair is
\begin{equation}
  \langle\mathbf{q},\mathbf{k}_i\rangle
  - \langle\mathbf{q},\hat{\mathbf{k}}_i\rangle
  = \langle\mathbf{q},\mathbf{r}_i\rangle.
\end{equation}
In expectation over random keys, $\mathbb{E}[\mathbf{r}_i]=0$ by
symmetry of the Lloyd-Max quantiser; but for a \emph{specific} key
$\mathbf{k}_i$ the value $\langle\mathbf{q},\mathbf{r}_i\rangle$ is
a fixed nonzero number. This is the per-instance bias that QJL
corrects.

One might ask whether this bias is immaterial in the softmax, since
it appears in both numerator and denominator:
\begin{equation}
  \frac{e^{a_i + b_i}}{\sum_j e^{a_j + b_j}},
  \quad b_i = \langle\mathbf{q},\mathbf{r}_i\rangle.
\end{equation}
If $b_i = b$ were constant across all keys, it would cancel exactly.
But $b_i$ is \emph{key-specific}: each key has its own quantisation
residual $\mathbf{r}_i$, so the biases do not cancel. A
high-attention key with large $|b_i|$ is unfairly re-weighted.

QJL corrects this key-specific bias. However, as shown in
Section~\ref{sec:kv_asymmetry}, the correction comes at the cost of
$2\pi$ times more variance in the inner product estimate, which
through Jensen's inequality creates more systematic softmax distortion
than the per-instance bias it removes.

\textbf{Implication for $V$.} The value tensor enters the output
\emph{linearly}: $\sum_i w_i \hat{\mathbf{v}}_i$. There is no
exponential, so variance is not amplified. The per-instance bias in
$V$ quantisation propagates directly and additively to the output,
making QJL's unbiased correction genuinely useful here without the
$2\pi$ variance penalty that makes it harmful for $K$.

\subsection{KL Divergence as Mechanistic Bridge}
\label{sec:kl_bridge}

Let $p_{\mathrm{ref}} = \mathrm{softmax}(QK^T/\sqrt{d})$ and
$p_{\mathrm{quant}} = \mathrm{softmax}(Q\hat{K}^T/\sqrt{d})$ denote
the reference and quantised attention weight distributions for a fixed
query. The KL divergence
\begin{equation}
  \mathrm{KL}(p_{\mathrm{ref}} \| p_{\mathrm{quant}})
  = \sum_i p_{\mathrm{ref},i} \log \frac{p_{\mathrm{ref},i}}{p_{\mathrm{quant},i}}
\end{equation}
is \emph{K-only by construction}: $V$ never enters the softmax, so $V$
quantisation has identically zero effect on this quantity. This makes KL
a clean mechanistic probe of the K-cache error path, decoupled from $V$.

The convexity of $e^x$ (Jensen's inequality) amplifies K direction errors
superlinearly into KL. For a key with score perturbation
$\epsilon_i = \langle\mathbf{q},\hat{\mathbf{k}}_i - \mathbf{k}_i\rangle$,
even zero-mean perturbations inflate softmax numerators:
$\mathbb{E}[e^{s_i+\epsilon_i}] = e^{s_i} \cdot e^{\sigma_i^2/2}$.
When $\sigma_i^2$ is non-uniform across keys — the generic case under any
quantisation scheme — the inflation is asymmetric and KL grows superlinearly
with the K direction error $\epsilon_K^{\mathrm{dir}}$.

This KL inflation has a direct routing consequence, measured by the
\emph{top-5 recall} $\mathrm{topk5} \in [0,1]$: the fraction of the
reference top-5 attended tokens recovered by the quantised distribution.
The routing error $1 - \mathrm{topk5}$ increases monotonically with KL,
and routing errors propagate to the output direction error $\epsilon_T^{\mathrm{dir}}$
through the weighted sum $\sum_i w_i \hat{\mathbf{v}}_i$.

\begin{figure*}[htbp]
  \centering
  \includegraphics[width=\textwidth]{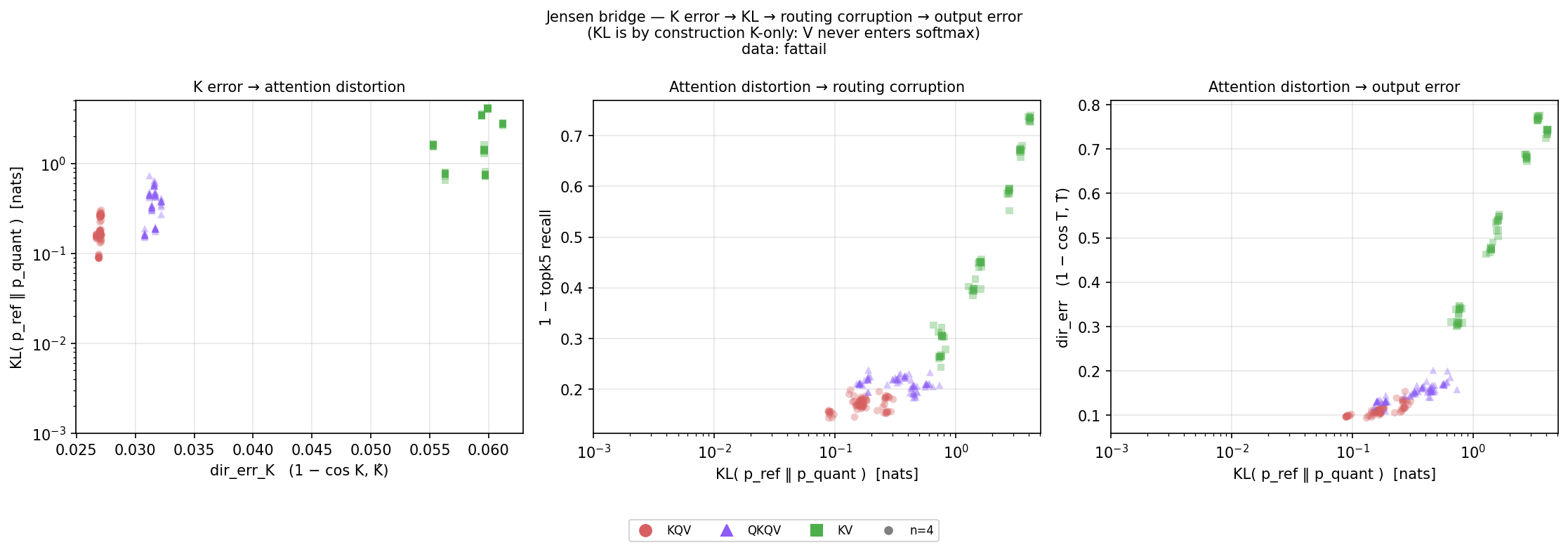}
  \caption{%
    \textbf{Jensen bridge} — the K-driven causal chain at budget $n=4$,
    fattail regime ($\nu=3$, log KL axis).
    Colour: KQV (red $\bullet$), QKQV (purple $\blacktriangle$),
    KV (green $\blacksquare$).
    \textbf{Left}: three separated clouds in
    $(\epsilon_K^{\mathrm{dir}},\,\mathrm{KL})$ space.
    KQV sits leftmost (WHT + $n$-bit scalar on $K$ minimises both);
    KV is far right (no WHT, heavy tails $\Rightarrow$ large K direction
    error, KL up to 4 nats).
    \textbf{Centre and right}: all three clouds collapse onto the same
    monotone curve in (KL, $1-\mathrm{topk5}$) and
    (KL, $\epsilon_T^{\mathrm{dir}}$) space.
    KL is the sufficient statistic: the scheme label is irrelevant once
    KL is observed. KL is K-only by construction; V enters the output
    linearly after the weights are fixed and is not shown.
  }
  \label{fig:jensen_bridge}
\end{figure*}

Figure~\ref{fig:jensen_bridge} shows this causal chain empirically at
budget $n=4$ on the fattail regime. The three scheme clouds separate
in panel~1 ($\epsilon_K^{\mathrm{dir}}$ vs.\ KL) but collapse onto
a single universal curve in panels~2 and 3 (KL vs.\ routing error,
KL vs.\ output error): \textbf{once KL is known, the scheme identity
disappears from all downstream metrics.} KL is the sufficient statistic
connecting K quantisation quality to routing corruption and output collapse.

\section{The K--V Asymmetry: Why QJL on Keys Harms Attention Quality}
\label{sec:kv_asymmetry}

TurboQuant applies QJL as a residual corrector to $V$ only.
A natural question is whether the same 1-bit sketch could be applied
to $K$ instead -- either as the sole quantisation or as an additive
corrector.  We show that under near-isotropic key distributions --
the regime where the WHT normalises marginals to
Beta (Section~\ref{lab:beta}) -- this is structurally incorrect:
WHT on $K$ is exactly invertible and preserves inner product quality,
whereas QJL on $K$ inflates inner product variance by at least $\pi/2$
and is amplified nonlinearly by the softmax.  For low-rank or
subspace-concentrated keys (Section~\ref{lab:beta}), the WHT does not
achieve its normalisation goal and the argument must be qualified, as
we discuss in Section~\ref{sec:qjl_key_experiment}.

\subsection{Invertibility: WHT vs.\ QJL}

Stage~1 applies the transform $H\,\mathrm{diag}(\mathbf{s})$, which is
\emph{exactly invertible}: since $\mathrm{diag}(\mathbf{s})^2 = I$ and
$H^T H = d\,I$, the original normalised vector is recovered as
\begin{equation}
  \hat{\mathbf{x}} = \frac{1}{d}\,\mathrm{diag}(\mathbf{s})\,H\,
                      \mathrm{dequant}\!\left(H\,\mathrm{diag}(\mathbf{s})
                      \,\hat{\mathbf{v}}\right).
\end{equation}
The only error in $\hat{\mathbf{k}}_i$ is scalar quantisation noise;
the transform itself introduces none.  The reconstructed key is
used in standard inner products $\langle\mathbf{q},\hat{\mathbf{k}}_i\rangle$
via any BLAS routine without modification.

QJL discards all per-coordinate magnitude information:
$\mathbf{q} = \mathrm{sign}(H\,\mathrm{diag}(\mathbf{s})\,\hat{\mathbf{v}})
\in \{-1,+1\}^d$.
No exact inversion exists.  The standard reconstruction
\begin{equation}
  \hat{\mathbf{r}} = \frac{\sqrt{\pi/2}\,\|\mathbf{r}\|}{d}\,
                     \mathrm{diag}(\mathbf{s})\,H\,\mathbf{q}
\end{equation}
is unbiased in expectation over the randomness of
$\mathrm{diag}(\mathbf{s})$, but for a \emph{fixed, stored}
sign vector $\mathbf{s}$ it is a fixed approximation with
irreducible per-instance error of order $\|\mathbf{r}\|/\sqrt{d}$
per coordinate.

\subsection{Inner Product Variance: a Unified 2\texorpdfstring{$\pi$}{pi} Argument}

Define the inner product error for a fixed query $\mathbf{q}$ and
key $\mathbf{k}$:
\begin{equation}
  \delta = \langle\mathbf{q}, \hat{\mathbf{k}}\rangle
         - \langle\mathbf{q}, \mathbf{k}\rangle.
\end{equation}
For \textbf{WHT + $B$-bit scalar} on $K$:
$\hat{\mathbf{k}} = \mathbf{k} + \mathbf{r}_B$ where
$\mathbf{r}_B$ is the scalar quantisation residual,
$\|\mathbf{r}_B\|^2 \approx \|\mathbf{k}\|^2\cdot\epsilon_B$
with $\epsilon_B$ the per-token relative MSE at $B$ bits.
\begin{equation}
  \mathrm{Var}[\delta]_{\mathrm{scalar}}
  = \frac{\|\mathbf{q}\|^2\|\mathbf{r}_B\|^2}{d}
  = \frac{\|\mathbf{q}\|^2\|\mathbf{k}\|^2\,\epsilon_B}{d}.
\end{equation}

For \textbf{QJL applied to $K$} (1-bit scheme, no scalar base):
the unbiased estimator has variance
\begin{equation}
  \mathrm{Var}[\delta]_{\mathrm{QJL\text{-}K}}
  = \frac{\pi}{2}\cdot\frac{\|\mathbf{q}\|^2\|\mathbf{k}\|^2}{d}.
\end{equation}

The ratio is
\begin{equation}
  \frac{\mathrm{Var}[\delta]_{\mathrm{QJL\text{-}K}}}
       {\mathrm{Var}[\delta]_{\mathrm{scalar}}}
  = \frac{\pi/2}{\epsilon_B}.
\end{equation}
Since $\epsilon_B < 1$ for any positive bit budget, QJL on $K$ is
always worse than scalar quantisation on $K$ by a factor of at least
$\pi/2 \approx 1.57$.  The factor grows as $\epsilon_B \to 0$
(higher bit budget), meaning QJL-on-$K$ becomes \emph{relatively
worse} as quality improves elsewhere.

\begin{observation}
Under equal bit budget, replacing scalar quantisation of $K$ with
a 1-bit QJL sketch inflates the inner product variance by a factor
$(\pi/2)/\epsilon_B \geq \pi/2$, independent of dimension $d$.
The inflated variance propagates through the softmax exponential
via Jensen's inequality, amplifying attention weight distortion
by the same factor.
This argument assumes WHT achieves its normalisation goal, so $\epsilon_B$
is the Beta-optimal scalar MSE.  For low-rank keys whose marginals are
not Beta after WHT, $\epsilon_B$ may be large; in that regime the scalar
path itself fails and the relevant remedy is a better rotation, not QJL.
\end{observation}

\subsection{The Non-Averaging Argument}
\label{sec:nonavg}

For $V$, the quantisation error enters the output \emph{linearly}:
$\sum_{i=1}^S w_i\,(\hat{\mathbf{v}}_i - \mathbf{v}_i)$.
Across a sequence of length $S$, errors with zero mean cancel at
rate $1/\sqrt{S}$.  This makes QJL's unbiasedness genuinely
valuable: the expected error is zero and variance shrinks with $S$.

For $K$, the error enters \emph{nonlinearly} through the softmax:
\begin{equation}
  w_i = \frac{e^{\langle\mathbf{q},\hat{\mathbf{k}}_i\rangle}}
             {\sum_j e^{\langle\mathbf{q},\hat{\mathbf{k}}_j\rangle}}.
\end{equation}
There is no averaging over $S$.  The softmax selects a
\emph{winner}: a single key $k^* = \arg\max_i
\langle\mathbf{q},\hat{\mathbf{k}}_i\rangle$ can receive
nearly all the weight.  If QJL's inner product estimate
places a different key at the top -- an event whose probability
grows with the inflated variance $(\pi/2)\epsilon_B^{-1}$ --
the entire output vector changes, not just one term.

This argmax-shift failure mode is empirically visible in the
focused-attention experiments:
when $\mathbf{q} \approx \mathbf{k}_{i^*}$ for a single dominant
key $i^*$, a small direction error in $\hat{\mathbf{k}}_{i^*}$
collapses the output to the wrong key.  QJL's larger inner product
variance makes this collapse more likely at every budget.

\subsection{Proposed Experiment: QJL Applied to Keys}
\label{sec:qjl_key_experiment}

To confirm the above analytically, we propose a controlled
ablation within the existing \texttt{pquant\_hip\_metrics} framework.
Three schemes are compared at equal total bit budget $B$:

\begin{enumerate}
  \item \textbf{Turbo (current)}: WHT+scalar at $B$ bits for $K$;
        WHT+scalar at $B-1$ bits + QJL for $V$.
  \item \textbf{QJL-K ablation}: QJL at $B$ bits for $K$
        (i.e.\ apply the sign sketch to the normalised key direction);
        WHT+scalar at $B$ bits for $V$ (no QJL on $V$).
  \item \textbf{Plain}: scalar at $B$ bits for both $K$ and $V$,
        no WHT, no QJL.
\end{enumerate}

The 6D error vector
\begin{equation}
  \mathbf{e} = ([\epsilon_K^{\mathrm{snr}},\epsilon_K^{\mathrm{dir}}],
         [\epsilon_V^{\mathrm{snr}},\epsilon_V^{\mathrm{dir}}],
         [\epsilon_T^{\mathrm{snr}},\epsilon_T^{\mathrm{dir}}])
\end{equation}

is measured across all four
key distributions (random, low-rank, fattail, focused) and
all five budgets $B \in \{2,3,4,5,6\}$.

\textbf{Predicted outcomes (near-isotropic inputs).}

\begin{itemize}
  \item $\epsilon_K^{\mathrm{snr}}$: QJL-K $\gg$ Turbo.
        Scalar WHT achieves Beta-optimal MSE; QJL replaces all
        coordinate precision with a single sign bit.

  \item $\epsilon_K^{\mathrm{dir}}$: not directly controlled by the
        $\pi/2$ argument, which concerns inner product
        \emph{variance}, not cosine distance.  For near-isotropic
        inputs, WHT+scalar preserves direction via exact inversion
        (Turbo $\leq$ QJL-K).  For low-rank inputs at low budgets,
        the WHT does not normalise the distribution; the scalar
        codebook then underperforms QJL's distribution-free
        direction estimate, and the ordering may reverse.

  \item $\epsilon_T^{\mathrm{dir}}$ (focused mode): QJL-K
        catastrophic regardless of key distribution.
        The argmax shift failure is triggered with high probability
        when attention is concentrated on a single key.
        The Mann-Whitney $r$ statistic is predicted to reach $r = +1$
        (QJL-K worst across the full distribution).

  \item $\epsilon_T^{\mathrm{dir}}$ (random mode): QJL-K moderately
        worse than Turbo for near-isotropic inputs.
        With diffuse attention the argmax shift is unlikely;
        the extra variance averages out across the sequence.

  \item $\epsilon_V^{\mathrm{snr}}$, $\epsilon_V^{\mathrm{dir}}$:
        QJL-K $\approx$ Turbo $\approx$ Plain (V is identical across
        schemes in this ablation).
\end{itemize}

The crossover between random and focused modes, and between isotropic
and low-rank key distributions, is the clearest empirical signature:
QJL-K fails at the softmax level (KL, focused mode) even when it may
improve direction recovery at low bit budgets for degenerate inputs.
This separates the inner product variance claim (always valid) from
the direction error claim (distribution- and budget-dependent).

\section{The Shannon--\texorpdfstring{$\varphi$}{phi} Plane}
\label{sec:framework}

Any vector reconstruction $\hat{\mathbf{x}} \approx \mathbf{x}$
admits a natural decomposition into two orthogonal error components.

\textbf{Scale error} (SNR): the per-token relative mean squared error
\begin{equation}
  \epsilon^{\mathrm{snr}} = \frac{\|\mathbf{x} - \hat{\mathbf{x}}\|^2}{\|\mathbf{x}\|^2}.
\end{equation}

\textbf{Direction error} ($\varphi$): the cosine distance
\begin{equation}
  \epsilon^{\mathrm{dir}} = 1 - \frac{\langle\mathbf{x},\hat{\mathbf{x}}\rangle}
                                      {\|\mathbf{x}\|\,\|\hat{\mathbf{x}}\|}.
\end{equation}
By construction $\epsilon^{\mathrm{dir}} \in [0, 2]$, with $0$
meaning perfect alignment and $2$ meaning anti-parallel.

Plotting $(\epsilon^{\mathrm{snr}}, \epsilon^{\mathrm{dir}})$ for a
population of trials defines the \emph{Shannon--$\varphi$
plane}~\cite{shannon1948mathematical}.  The two axes are not chosen
for presentational convenience: they are the canonical Euclidean and
angular measures of vector reconstruction quality, standard in signal
processing and information geometry.  They are \emph{orthogonal} -- a
scheme can achieve perfect direction with poor scale (e.g.\ a
unit-normalised reconstruction), or perfect scale with poor direction
-- so neither axis subsumes the other.  This orthogonality is what
motivates plotting them jointly rather than collapsing to a single
scalar metric. We recall that we start with the hypothesis that we
have a vector on a unit sphere, any quantization will affect where in
the sphere the new vector will land.

\subsection{Theoretical Bounds on Both Axes}

The measurement space is not unbounded: both axes admit theoretical
limits derived from quantisation theory and the Johnson-Lindenstrauss
lemma~\cite{johnson1984extensions}, making them \emph{physical} measures rather than opportunistic
ones.

\textbf{Direction bound.} For any 1-bit JL sketch of a
$d$-dimensional vector the variance of the unbiased direction
estimator per coordinate is $O(1/d)$, giving an asymptotic floor
$\epsilon^{\mathrm{dir}} \geq C/\sqrt{d}$ for a universal constant
$C$.  At $d = 128$ this is non-trivial: no 1-bit sketch can do better,
regardless of how the bits are arranged.

\textbf{Scale bound.} The Lloyd-Max quantiser minimises MSE for a
given distribution and bit budget $B$.  For the
Beta$\!\left(\frac{d-1}{2},\frac{d-1}{2}\right)$ marginals produced
by WHT, the per-bit SNR gain follows the $6\,\mathrm{dB/bit}$ rule
asymptotically: $\epsilon^{\mathrm{snr}}_B \approx \epsilon^{\mathrm{snr}}_0
\cdot 4^{-B}$.  No scalar quantiser for this distribution achieves
lower MSE at the same budget.

These bounds define the walls of the measurement space.  Deviations
from the theoretical floors measure the cost of design choices, not
measurement artefacts.

\subsection{Why KL Divergence Is Insufficient}

The comparison in Section~\ref{sec:experiments} uses KL divergence
between the true and quantised softmax attention distributions.
KL is a natural end-to-end metric, but it conflates four distinct
quantities: scale error in $K$, direction error in $K$, scale error
in $V$, and direction error in $V$.  A scheme that destroys $K$
direction while improving $V$ scale can show neutral KL divergence
while failing catastrophically on specific queries.

Moreover, KL is an expectation and is dominated by typical cases,
masking rare but operationally severe events -- most notably the
argmax-shift failure of Section~\ref{sec:nonavg}, which occurs when a
mis-quantised dominant key shifts the softmax winner.  The
Shannon--$\varphi$ plane exposes this failure mode directly in the
$(\epsilon_K^{\mathrm{snr}}, \epsilon_K^{\mathrm{dir}})$ quadrant,
invisible to any scalar metric.

A subtler limitation is that $K$ and $V$ errors can be
\emph{anticorrelated} in the output $T$.  A shift in attention weights
caused by $K$ quantisation error may happen to select $V$ vectors
closer to the true weighted sum, partially or fully cancelling the
$K$ error in the output.  KL divergence, measured at $T$, cannot
distinguish this cancellation from genuine accuracy: a scheme with
large $K$ error and fortuitous $V$ cancellation looks identical to
one with small errors in both.  Only a decomposition that measures
$K$ and $V$ independently -- as the 6D vector does -- can reveal
whether agreement at $T$ reflects quality or cancellation.

\subsection{The 6D Error Vector}

We extend the Shannon--$\varphi$ plane to three measurement sites:
the $K$ cache, the $V$ cache, and the attention output
$T = \mathrm{softmax}(QK^T/\!\sqrt{d})\,V$.  Each site contributes
a pair $(\epsilon^{\mathrm{snr}}, \epsilon^{\mathrm{dir}})$,
yielding a six-dimensional error vector per trial:
\begin{equation}
  \mathbf{e} = \bigl(
    \epsilon_K^{\mathrm{snr}},\,\epsilon_K^{\mathrm{dir}},\,
    \epsilon_V^{\mathrm{snr}},\,\epsilon_V^{\mathrm{dir}},\,
    \epsilon_T^{\mathrm{snr}},\,\epsilon_T^{\mathrm{dir}}
  \bigr).
  \label{eq:6d}
\end{equation}
The Euclidean distances $d_K$, $d_V$, $d_T$ within each
Shannon--$\varphi$ plane aggregate each site into a scalar for
ranking, while preserving the directional decomposition for
attribution.  The full vector $\mathbf{e}$ is used for
distribution-level comparisons via the energy distance.

\subsection{Statistical Instruments and the Quorum Principle}

For each pairwise comparison (per budget $B$) three instruments are
applied to the empirical distributions of $\mathbf{e}$ (i.e., error
vector):

\textbf{Mann-Whitney $U$} (1D ranked): the rank-biserial
correlation $r \in [-1, +1]$ for each of $d_K$, $d_V$, $d_T$.
Convention: negative $r$ = scheme A better; positive $r$ = scheme B
better (scheme A and B are defined per comparison).
Nonparametric: no distributional assumption required.

\textbf{Kolmogorov-Smirnov}: maximum CDF separation $D$ and
$p$-value for each scalar projection.  Sensitive to distributional
shape, not only central tendency.

\textbf{Energy distance}~\cite{szekely2013energy} (6D):
\begin{equation}
  E(X,Y) = 2\,\mathbb{E}[\|\mathbf{x}-\mathbf{y}\|]
           - \mathbb{E}[\|\mathbf{x}-\mathbf{x}'\|]
           - \mathbb{E}[\|\mathbf{y}-\mathbf{y}'\|],
\end{equation}
evaluated in the Shannon coordinate system (raw dimensions, equally
weighted) and the Lloyd-Max (LM) system (scale dimensions amplified by
the scalar loss factor $\approx 5.44$, above the Lloyd-Max SNR
threshold).  Significance is assessed by a permutation test with the
Phipson-Smyth correction~\cite{phipson2010permutation}. Here we can
plug in any further parametric and non parametric distance, we choose
one where a confidence level can be estimated from the original
sample.

Three instruments in two coordinate systems constitute a
\emph{quorum}: a conclusion requires agreement from at least two
instruments.  Disagreement between coordinate systems is itself
informative -- it identifies effects that are scale-specific rather
than universal.

The intent is to give a quantitative measure to a natural question: if
we have two quantisations and we cannot distinguish their numerical
properties in a general scenario, which one is worth to deploy will be
related to other characteristics. Another intent is to show that K and
V properties are asymmetric and they can interfere and we should build
a more complex model of their effects.

\begin{figure}[t]
  \centering
  \begin{subfigure}[t]{0.48\textwidth}
    \centering
    \includegraphics[width=\textwidth]{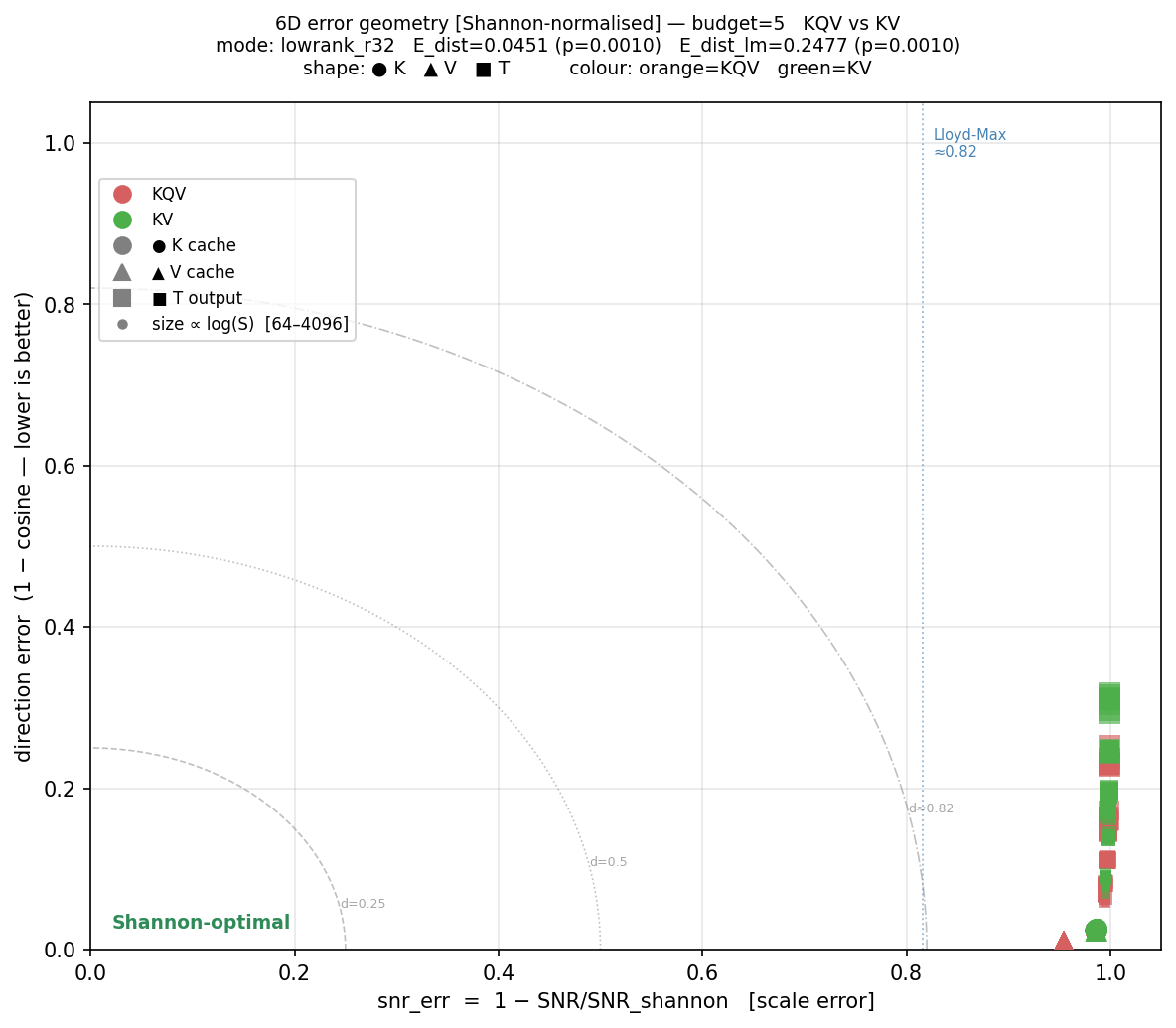}
    \caption{Low-rank rank=32, $B=5$. Scale error is
      resolution-limited for both schemes; the separation is in
      direction: KQV T (orange $\blacksquare$) at
      $\epsilon_T^{\mathrm{dir}} = 0.111$ vs.\ KV at $0.160$.
      MW $r=+0.958$, $E_{6d}=0.047$.}
    \label{fig:lowrank_r32_B5}
  \end{subfigure}
  \hfill
  \begin{subfigure}[t]{0.48\textwidth}
    \centering
    \includegraphics[width=\textwidth]{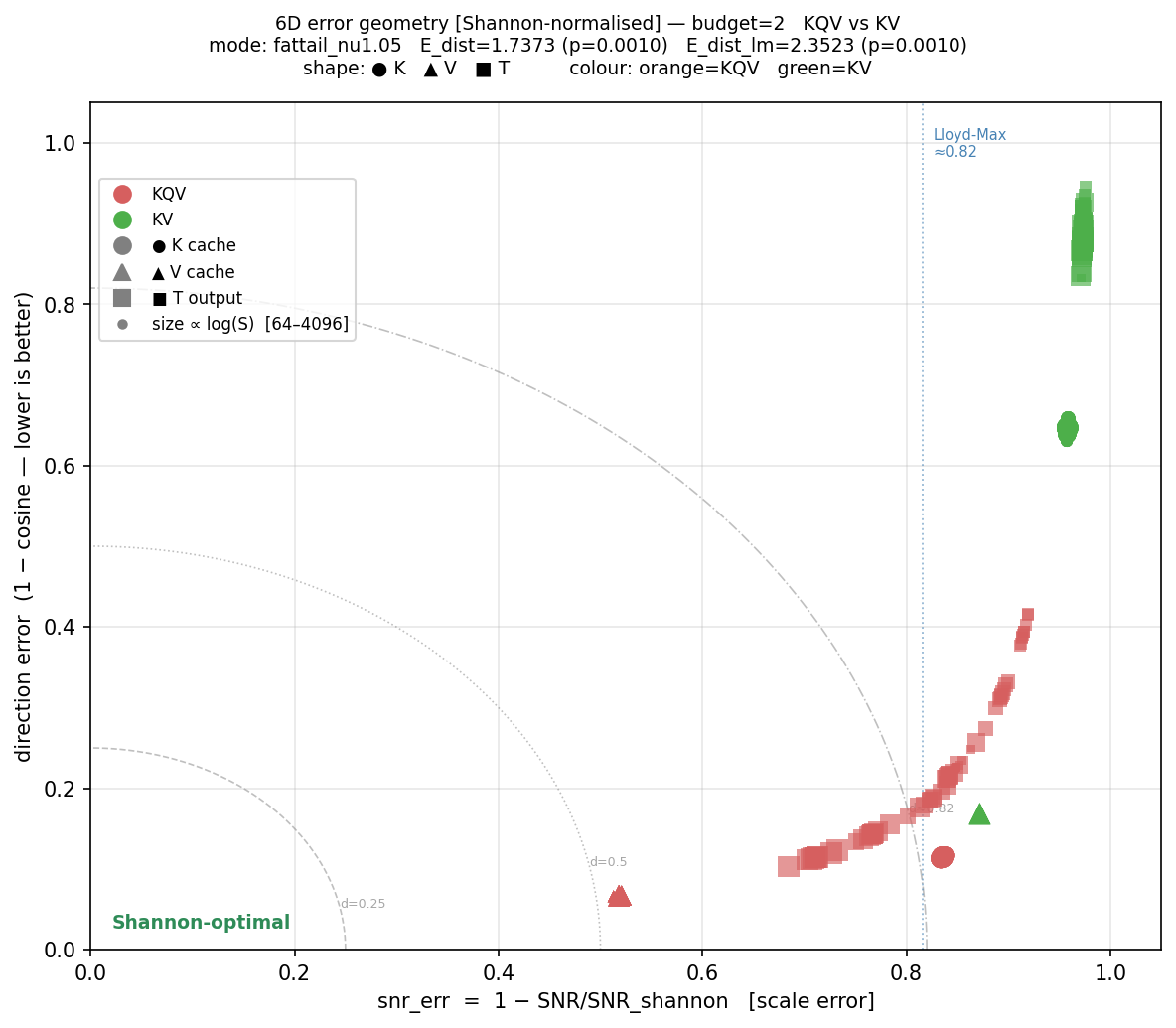}
    \caption{Fattail $\nu=1.05$, $B=2$ (near-Cauchy, coarsest budget).
      KV K (green $\bullet$, top-right): full codebook saturation
      ($\epsilon_K^{\mathrm{dir}}=0.645$). KQV K (orange $\bullet$):
      WHT maps input to $\pm1/\!\sqrt{d}$ before quantisation
      ($\epsilon_K^{\mathrm{dir}}=0.115$). KV T trapped at
      $\epsilon_T^{\mathrm{dir}}\approx 0.89$.
      $E_{6d}=1.710$, MW $r=+1.000$.}
    \label{fig:fattail_nu105_B2}
  \end{subfigure}
  \caption{Shannon--$\varphi$ plane comparisons (orange = KQV,
    green = KV; $\bullet$ K cache, $\blacktriangle$ V cache,
    $\blacksquare$ T output). Left: the near-Gaussian low-rank regime
    where the statistical framework is needed to detect the difference.
    Right: the extreme fat-tail regime where visual inspection suffices.}
  \label{fig:combined_geometry}
\end{figure}

\begin{figure}[t]
  \centering
  \includegraphics[width=0.65\textwidth]{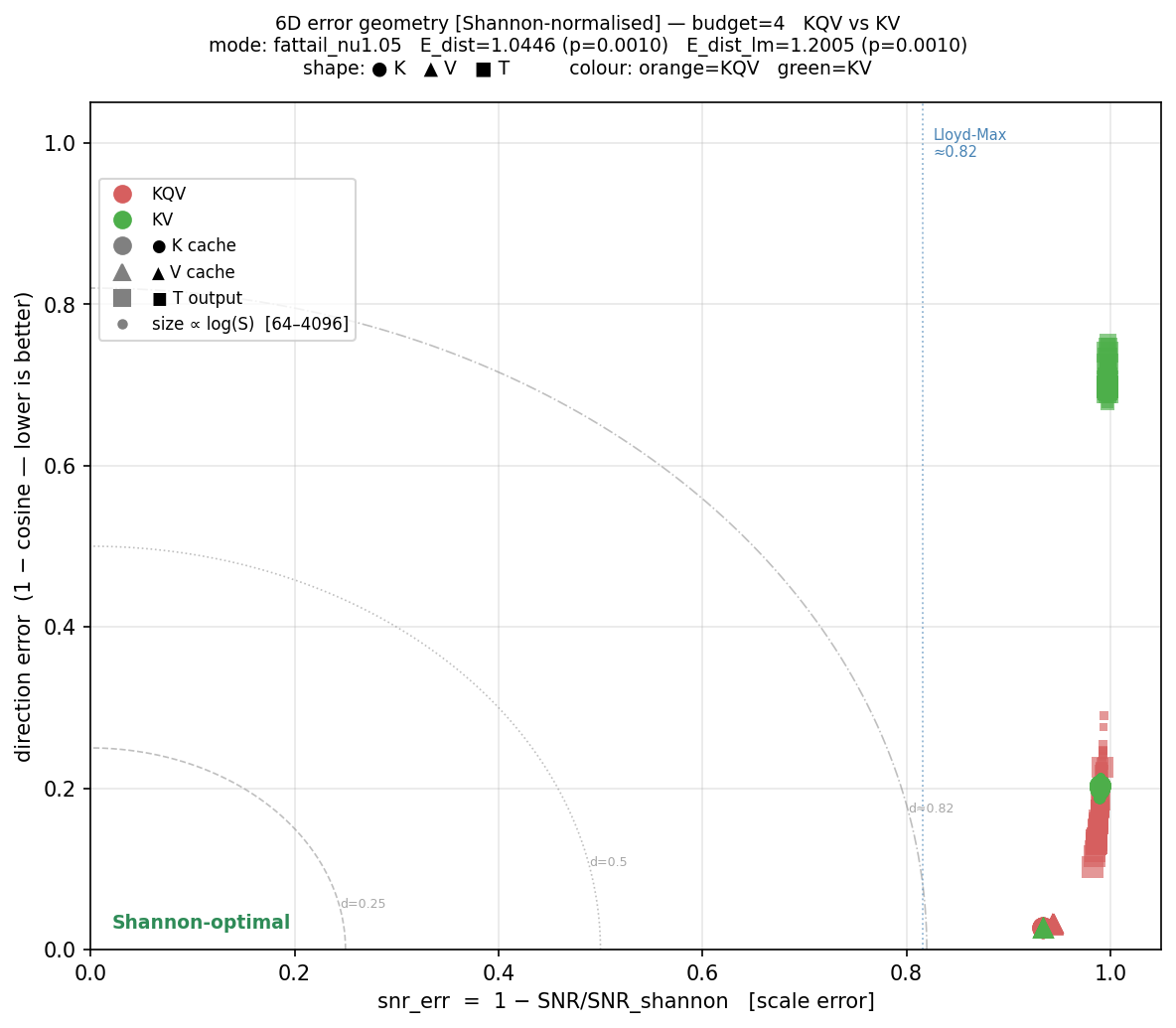}
  \caption{Same mode ($\nu=1.05$) at $B=4$.  With two additional bits
           KQV K (orange $\bullet$) remains stable at
           $\epsilon_K^{\mathrm{dir}} = 0.027$; KV K retreats from
           $0.645$ to $0.202$ but the gap persists.
           Both V caches move toward the Lloyd-Max line as bit budget
           increases.  The geometric separation shrinks but the
           ordering is unchanged: MW $r=+1.000$, energy distance $1.004$.}
  \label{fig:fattail_nu105_B4}
\end{figure}

In Figure \ref{fig:fattail_nu105_B2} and \ref{fig:fattail_nu105_B4},
we show examples where this geometrical representation is striking and
there is no need for further investigations than a visual
inspection. The statistical inference helps when the scenario is not
so clear such as in Figure \ref{fig:lowrank_r32_B5}.

\section{Experiments: Fair Budget Comparison}
\label{sec:experiments}

Section~\ref{sec:kv_asymmetry} predicts that QJL applied to $K$ is
structurally worse than scalar quantisation under any fair bit budget.
The simplest way to verify this is with KL divergence between the
true and quantised softmax distributions --- a single, well-understood
number that any reader can interpret directly.

\subsection{Setup}
We compare:
\begin{itemize}
  \item \textbf{MSE}: Lloyd-Max at $B$ bits per coordinate.
  \item \textbf{MSE+QJL}: Lloyd-Max at $(B-1)$ bits + 1-bit QJL sketch
        (\emph{equal total bit budget}).
\end{itemize}
Three key distributions are tested: \texttt{gaussian} ($K \sim
\mathcal{N}(0,I)$), \texttt{heavy\_tail} (10\% of keys scaled by
$10\times$), and \texttt{low\_rank} ($K = AB$ with rank $d/8$).

Why we choose these distributions. The Gaussian and the Beta should be
asymptotically indistinguishable, providing a clean reference.
Heavy-tailed keys expose the gap between the Beta support $[-1,1]$
and the Gaussian approximation after normalization and rotation.
Low-rank keys mimic clustered embeddings: any rotation preserves
the subspace structure, breaking the isotropy assumption and
rendering the marginal Beta guarantee insufficient for joint
quantisation quality.

Performance is measured by KL divergence between true and quantised
softmax attention distributions, over 100 trials.

\subsection{Main Result when applying softmax}
\begin{observation}
Under equal bit budget, \textbf{MSE alone outperforms MSE+QJL} for all
distributions at $d \in \{2,8,128\}$, and for gaussian and low\_rank
at $d=1024$. Heavy\_tail flips to MSE+QJL at $d=1024$.
\end{observation}

The explanation is quantitative. The correct comparison is between
inner product error variances, since it is inner products that enter
the softmax.

\textbf{B-bit MSE only.}
Each additional bit in a Lloyd-Max quantiser roughly halves the MSE
(6\,dB per bit), so removing one bit quadruples the residual energy:
\begin{equation}
  \|\mathbf{r}_{B-1}\|^2 \approx 4\|\mathbf{r}_B\|^2.
\end{equation}
The inner product error $\langle\mathbf{q},\mathbf{r}_B\rangle$
has variance
\begin{equation}
  \sigma^2_{\mathrm{MSE}} = \frac{\|\mathbf{r}_B\|^2}{d}.
\end{equation}

\textbf{(B-1)-bit MSE + QJL.}
The 1-bit QJL sketch estimates
$\langle\mathbf{q},\mathbf{r}_{B-1}\rangle$ with variance
\begin{equation}
  \sigma^2_{\mathrm{QJL}}
  = \frac{\pi}{2}\cdot\frac{\|\mathbf{r}_{B-1}\|^2}{d}
  = \frac{\pi}{2}\cdot\frac{4\|\mathbf{r}_B\|^2}{d}
  = \frac{2\pi\,\|\mathbf{r}_B\|^2}{d}.
\end{equation}

\textbf{Conclusion.}
The ratio of the two inner product error variances is
\begin{equation}
  \frac{\sigma^2_{\mathrm{QJL}}}{\sigma^2_{\mathrm{MSE}}}
  = 2\pi \approx 6.28,
\end{equation}
independent of $d$. QJL always loses by a constant factor of $2\pi$
under fair budget. \textbf{No dimension makes QJL worth the bit
trade.}

\begin{figure*}[htbp]
  \centering
  \includegraphics[width=\textwidth]{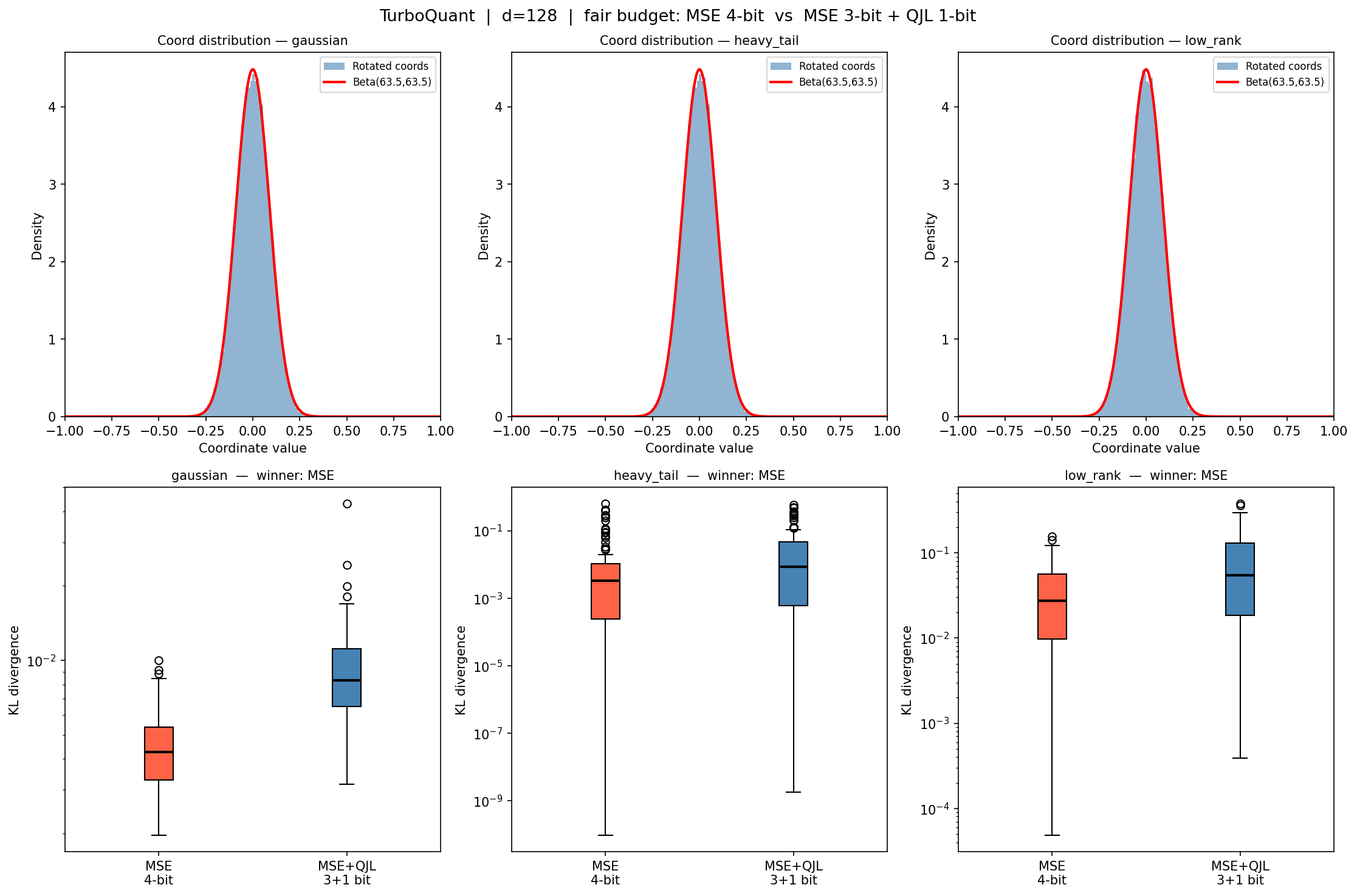}
  \caption{%
    $d=128$, Beta$(63.5,63.5)$ — deeply concentrated, nearly Gaussian.
    \textit{Top}: all three distributions match the Beta curve closely;
    the subspace structure of \texttt{low\_rank} is already hidden in
    the marginal histogram.
    \textit{Bottom}: MSE wins across all distributions under equal
    budget, consistent with the $2\pi$ variance penalty of QJL.
    Circles denote outliers beyond $1.5\times\mathrm{IQR}$
    (${\approx}2.7\sigma$ for a Gaussian).
  }
  \label{fig:d128}
\end{figure*}

Figure~\ref{fig:d128} shows results at $d=128$, one of the most
common head dimensions in deployed transformers. MSE wins across all
distributions, and the low\_rank histogram is already indistinguishable
from Beta --- yet the KL divergence gap between distributions persists,
foreshadowing the catastrophic failure at $d=1024$.

The heavy\_tail exception at $d=1024$ arises because distribution
mismatch creates a non-negligible bias in Stage~1 that QJL can correct,
and the $1/d$ variance suppression is large enough to make the trade
worthwhile.

\textbf{Conclusion}: QJL is justified only as \emph{additive overhead},
not as a replacement for MSE bits.

\section{Factorial Design and Experimental Results}
\label{sec:factorial}

The KL result above confirms the main claim at the softmax level.
But as noted in Section~\ref{sec:framework}, $K$ and $V$ errors can
be anticorrelated in the output $T$: a scheme with inflated $K$ error
may appear competitive in KL if $V$ errors happen to compensate.
To see whether the KL result reflects genuine quality or cancellation,
and to identify which component of TurboQuant is responsible for what,
we decompose the error across all six dimensions.

\subsection{Direct Confirmation of K--V Asymmetry}
\label{sec:kv_direct}

Using the scheme names introduced in Section~\ref{sec:schemes},
we compare \textbf{KQV} against \textbf{QKQV} at equal effective bit
budget $K_{\mathrm{eff}} = V_{\mathrm{eff}} = n$.  The $V$ cache is
identical between the two schemes; all differences are attributable
to $K$ alone (see Table~\ref{tab:schemes}).

Table~\ref{tab:kv_asymmetry} reports median $\epsilon_K^{\mathrm{snr}}$,
$\epsilon_K^{\mathrm{dir}}$, and $\epsilon_T^{\mathrm{dir}}$ together
with the Mann-Whitney $r$ for the 6D distance $d_6$, across all four
distributions and budgets $n \in \{2,3,4,5,6\}$.

\begin{table*}[htbp]
\centering
\small
\caption{KQV vs.\ QKQV: median geometric $K$ errors and output
direction error per distribution and budget.
$V$ is identical between the two schemes; all differences are on $K$.
\textbf{Bold} marks the winner on geometric $K$ quality
($\epsilon_K^{\mathrm{snr}}$ and $\epsilon_K^{\mathrm{dir}}$ always agree).
$^*$ denotes $p > 0.05$ (energy-distance permutation test, 999 permutations).
Lower is better for all $\epsilon$ columns.}
\label{tab:kv_asymmetry}
\begin{tabular}{lrccccccr}
\toprule
& & \multicolumn{2}{c}{$\epsilon_K^{\mathrm{snr}}$}
  & \multicolumn{2}{c}{$\epsilon_K^{\mathrm{dir}}$}
  & \multicolumn{2}{c}{$\epsilon_T^{\mathrm{dir}}$}
  & MW $r$ \\
\cmidrule(lr){3-4}\cmidrule(lr){5-6}\cmidrule(lr){7-8}
Dist. & $n$ & KQV & QKQV & KQV & QKQV & KQV & QKQV & ($d_6$) \\
\midrule
\texttt{lowrank}
  & 2 & 0.872 & \textbf{0.519} & 0.168 & \textbf{0.067} & 0.120 & \textbf{0.101} & $-1.000$ \\
  & 3 & 0.919 & \textbf{0.911} & 0.077 & \textbf{0.075} & 0.145 &          0.148  & $-0.449$ \\
  & 4 & \textbf{0.934} & 0.944 & \textbf{0.027} & 0.032 & \textbf{0.075} & 0.076 & $+0.627$ \\
  & 5 & 0.986 & \textbf{0.954} & 0.024 & \textbf{0.010} & 0.093 & \textbf{0.027} & $-1.000$ \\
  & 6 & \textbf{0.966} & 0.984 & \textbf{0.003} & 0.006 & \textbf{0.019} & 0.025 & $+0.993$ \\
\midrule
\texttt{focused}
  & 2 & 0.872 & \textbf{0.519} & 0.168 & \textbf{0.067} & 0.067 &          0.067  & $-1.000$ \\
  & 3 & 0.919 & \textbf{0.912} & 0.078 & \textbf{0.075} & 0.075 &          0.076  & $-0.977$ \\
  & 4 & \textbf{0.934} & 0.944 & \textbf{0.027} & 0.032 & 0.032 &          0.032  & $+1.000$ \\
  & 5 & 0.987 & \textbf{0.954} & 0.026 & \textbf{0.010} & 0.010 &          0.010  & $-1.000$ \\
  & 6 & \textbf{0.968} & 0.982 & \textbf{0.004} & 0.006 & 0.006 &          0.006  & $+0.994$ \\
\midrule
\texttt{fattail}
  & 2 & 0.868 & \textbf{0.504} & 0.162 & \textbf{0.065} & 0.306 & \textbf{0.189} & $-1.000$ \\
  & 3 & 0.919 & \textbf{0.910} & 0.076 & \textbf{0.073} & 0.248 & \textbf{0.231} & $-0.774$ \\
  & 4 & \textbf{0.934} & 0.944 & \textbf{0.027} & 0.032 & \textbf{0.102} & 0.138 & $+0.984$ \\
  & 5 & 0.983 & \textbf{0.955} & 0.021 & \textbf{0.010} & 0.073 & \textbf{0.048} & $-1.000$ \\
  & 6 & \textbf{0.966} & 0.978 & \textbf{0.003} & 0.005 & \textbf{0.019} & 0.023 & $+1.000$ \\
\midrule
\texttt{random}
  & 2 & 0.872 & \textbf{0.519} & 0.169 & \textbf{0.067} & 0.202 & \textbf{0.121} & $-1.000$ \\
  & 3 & 0.919 & \textbf{0.911} & 0.078 & \textbf{0.075} & 0.489 &    0.548$^*$    & $-0.069^*$ \\
  & 4 & \textbf{0.934} & 0.944 & \textbf{0.027} & 0.032 & 0.354 &    0.435$^*$    & $+0.158^*$ \\
  & 5 & 0.987 & \textbf{0.954} & 0.024 & \textbf{0.010} & 0.246 & \textbf{0.233} & $-0.353$ \\
  & 6 & \textbf{0.969} & 0.980 & \textbf{0.004} & 0.006 & 0.161 &    0.155$^*$    & $+0.220^*$ \\
\bottomrule
\end{tabular}
\end{table*}

Three findings stand out.

\textbf{Geometric $K$ quality: a budget-dependent crossover.}
The winner on $(\epsilon_K^{\mathrm{snr}}, \epsilon_K^{\mathrm{dir}})$
follows the identical pattern across all four distributions:
QKQV wins at $n \in \{2, 3\}$ (and mostly at $n=5$), KQV wins at
$n \in \{4, 6\}$.  Both axes always agree.  For \texttt{random}
(near-isotropic) inputs the differences at $n = 3, 4, 6$ are not
significant ($p > 0.05$): when the WHT achieves its normalisation
goal, trading a scalar bit for a QJL correction is geometrically
neutral.  For all other distributions the crossover is highly
significant and consistent.  Rank and low-rank sweeps (not shown in
this table) confirm the pattern is invariant to both rank and tail
weight --- it is a property of the bit-arithmetic, not the
distribution.

\textbf{At $n=4$, KQV wins on every measure.}
At the practically dominant 4-bit budget, KQV achieves better
geometric $K$ reconstruction, lower KL divergence, and better overall
$d_6$ across all four distributions, all ranks, and all tail weights
tested.  The geometric separation is visible by inspection of the
Shannon--$\varphi$ plane.  This is the most operationally relevant
finding: at the budget that matters for deployment, KQV is
unambiguously superior.

\textbf{KL and routing track the alternating pattern.}
Table~\ref{tab:kl_topk5} shows that KL divergence and top-5 routing
recall (Section~\ref{sec:kl_bridge}) agree with $d_6$ on the winner
at every budget. The scheme with lower KL always has lower routing
error and lower $\epsilon_T^{\mathrm{dir}}$, with one marginal
exception at $n=3$ where all effect sizes are small. At $n=4$ the
alignment is unambiguous: KQV achieves median KL $= 0.167$ vs.\
QKQV's $0.377$ ($2.25\times$ lower), and top-5 recall $0.831$ vs.\
$0.789$ (MW $r = -0.983$, $p < 10^{-23}$). The Jensen mechanism of
Section~\ref{sec:kl_bridge} explains this: QKQV's QJL on $K$
inflates K direction error, which the softmax exponential amplifies
superlinearly into KL, corrupting routing and cascading into output
collapse. The scheme with lower $\epsilon_K^{\mathrm{dir}}$ produces
lower KL, which produces better routing, which produces lower
$\epsilon_T^{\mathrm{dir}}$ — the causal chain is fully traceable.

\begin{table}[h]
\centering
\small
\caption{QKQV vs.\ KQV on fattail ($\nu=3$): MW rank-biserial $r$
  for $d_6$, KL divergence, and routing error $1-\mathrm{topk5}$,
  with median KL per scheme.
  Sign convention: $r > 0$ means QKQV lower (QKQV wins);
  $r < 0$ means KQV lower (KQV wins). All $p < 0.005$ except $\dagger$.}
\label{tab:kl_topk5}
\begin{tabular}{rrrrrr}
\toprule
$n$ & $r_{d_6}$ & $r_{\mathrm{KL}}$ & KL$_{\mathrm{QKQV}}$ & KL$_{\mathrm{KQV}}$ & $r_{1-\mathrm{topk5}}$ \\
\midrule
2 & $+1.000$ & $+1.000$ & 0.318 & 0.905 & $+0.869$ \\
3 & $+0.911$ & $+0.284^\dagger$ & 0.453 & 0.497 & $-0.397$ \\
4 & $-0.989$ & $-0.742$ & 0.377 & \textbf{0.167} & $-0.983$ \\
5 & $+1.000$ & $+0.709$ & 0.067 & 0.134 & $+0.992$ \\
6 & $-1.000$ & $-0.762$ & 0.029 & \textbf{0.015} & $-0.970$ \\
7 & $+1.000$ & $+0.778$ & 0.007 & 0.013 & $+0.804$ \\
\bottomrule
\end{tabular}
\end{table}

\textbf{Geometric $K$ quality $\neq$ softmax $K$ quality.}
Despite winning the geometric $K$ metrics at $n \in \{2, 3\}$,
QKQV produces consistently higher KL divergence than KQV at
every budget and distribution.  The $\pi/2$ inner product variance
inflation (Section~\ref{sec:kv_asymmetry}) harms the softmax
independently of how well the $K$ vector is reconstructed geometrically.
These are two genuinely different quantities, and the 6D framework
exposes both.

\textbf{Focused mode: $K$ errors wash out.}
In the \texttt{focused} distribution, $\epsilon_T^{\mathrm{dir}}$
is nearly identical for both schemes at $n \geq 3$.
Concentrated attention averages out $K$ reconstruction differences;
only the inner product variance (captured by KL, not by $d_6$)
discriminates the two schemes in this regime.

The remainder of this section uses the \textbf{KQV vs KV} comparison
to isolate the individual contributions of WHT and QJL-on-$V$.
The four distributions activate different subsets of the mechanisms:

\begin{center}
\begin{tabular}{lcc}
\toprule
Mode & WHT benefit & QJL-on-$V$ benefit \\
\midrule
\texttt{random}   & None (already isotropic) & Yes \\
\texttt{low-rank} & Scale / energy spreading & Yes \\
\texttt{fattail}  & Scale + direction        & Yes \\
\texttt{focused}  & Partial                  & Yes (but masked) \\
\bottomrule
\end{tabular}
\end{center}

\texttt{random} (isotropic Gaussian) isolates QJL-on-$V$: WHT has no
effect on a distribution already uniform on the sphere, so any
difference between KQV and KV at this mode is attributable solely
to the QJL residual corrector on $V$.  The contrast
$\texttt{fattail} - \texttt{random}$ isolates the WHT direction
benefit on $K$.  \texttt{low-rank} additionally activates WHT's
energy-spreading along the scale axis.  \texttt{focused}
($\mathbf{q} \approx \mathbf{k}_{i^*} + \boldsymbol{\eta}$ for a
single dominant key $i^*$) stress-tests the argmax-shift failure mode
predicted in Section~\ref{sec:kv_asymmetry}.

\subsection{QJL on V: a Rank-Invariant Win}

Across all four modes and all budgets $B \in \{2,3,4,5,6\}$, the
Mann-Whitney $r$ for $d_V$ is $-1.000$: KQV wins the $V$ direction
error perfectly at every distribution and every rank compared to KV.
This is the empirical realisation of the JL unbiasedness guarantee ---
the sketch is unconditionally unbiased regardless of the intrinsic
dimensionality of the key subspace.

A non-monotone exception is instructive: at $B \in \{4,6\}$ bits KV
wins $d_V$ over KQV.  The QJL binary residual does not compensate
for the lost scalar bit at those specific depths.  The crossover is
perfectly consistent across all four modes --- it is a property of the
bit-depth arithmetic, not of the distribution.

\subsection{Low-Rank Sweep: a Joint, Not Marginal, Failure}

We sweep intrinsic rank $\in \{1,2,4,8,16,32,64\}$ across all five
budgets.  The first finding is that \emph{both} KQV and KV
maintain approximately constant $K$ reconstruction quality across
the full rank range:

\begin{center}
\begin{tabular}{rcccc}
\toprule
Rank & \multicolumn{2}{c}{KQV} & \multicolumn{2}{c}{KV} \\
     & $\epsilon_K^{\mathrm{snr}}$ & $\epsilon_K^{\mathrm{dir}}$
     & $\epsilon_K^{\mathrm{snr}}$ & $\epsilon_K^{\mathrm{dir}}$ \\
\midrule
 1 & 0.9353 & 0.0279 & 0.9331 & 0.0269 \\
 8 & 0.9346 & 0.0272 & 0.9346 & 0.0272 \\
64 & 0.9343 & 0.0272 & 0.9341 & 0.0271 \\
\bottomrule
\end{tabular}
\end{center}

This is the marginal guarantee in action: WHT makes individual
coordinates Beta-distributed for one scheme but the dimension factor
is large enough that the other converge to a very similar Gaussian
without long tail, so the Lloyd-Max codebook works equally well for
both regardless of rank (on average).  Unlike the $\nu$ sweep, where
KV's $\epsilon_K^{\mathrm{dir}}$ grows $7\times$ as tails grow
heavier, here both schemes hold flat.  The rank failure mode is
\emph{joint}, not marginal.

The joint failure appears in the output $T$.  KQV's direction
error on $T$ grows with rank but remains bounded; KV's diverges:

\begin{center}
\begin{tabular}{rccc}
\toprule
Rank & KQV $\epsilon_T^{\mathrm{dir}}$ & KV $\epsilon_T^{\mathrm{dir}}$ & MW $r$ \\
\midrule
 1 & 0.068 & 0.303 & $+1.000$ \\
 8 & 0.120 & 0.325 & $+1.000$ \\
32 & 0.169 & 0.499 & $+1.000$ \\
64 & 0.182 & 0.605 & $+1.000$ \\
\bottomrule
\end{tabular}
\end{center}

The concentrated subspace structure creates peaked softmax
distributions; small errors on the dominant keys are amplified
exponentially.  WHT's energy-spreading limits this amplification;
KV's codebook, applied to the raw joint-dependent structure,
does not.

\textbf{Budget non-monotone.}  A second finding is that the KQV
advantage is non-monotone in budget:

\begin{center}
\begin{tabular}{rcl}
\toprule
Budget & MW $r$ (rank=32) & \\
\midrule
2 & $+1.000$ & KQV wins strongly \\
3 & $+0.700$ & KQV wins \\
4 & $+0.145$ & neutral \\
5 & $+0.958$ & KQV wins \\
6 & $-0.923$ & \textbf{KV wins} \\
\bottomrule
\end{tabular}
\end{center}

At $B=6$, KV wins across all ranks.  At this budget KV uses a
full 6-bit scalar on $V$; KQV uses 5-bit scalar~+~QJL, and the
extra QJL bit does not compensate for the lost scalar precision.
The $K$ metrics are essentially identical at $B=6$ ($\epsilon_K^{\mathrm{dir}}
\approx 0.003$ for both), so $V$ decides --- and KV's undivided
6-bit budget wins.  This is the QJL non-monotone behaviour made
visible in a near-Gaussian setting where the $K$ effect is neutral.

\subsection{WHT on Fat-Tail K: Direction Correction}

In \texttt{fattail} mode (component-wise $t_\nu$, $\nu = 3$), WHT
spreads the outlier energy uniformly across all $d$ coefficients,
correcting the component-wise concentration that would otherwise cause
specific coordinates to dominate inner products.  The result is a
direction benefit on $K$ absent in \texttt{random} mode.  The
difference $\epsilon_K^{\mathrm{dir}}(\texttt{fattail}) -
\epsilon_K^{\mathrm{dir}}(\texttt{random})$, evaluated per budget,
isolates the pure WHT direction contribution, decoupled from QJL.
KQV wins $d_T$ across all budgets in this mode: the WHT direction
benefit on $K$ overwhelms any QJL non-monotone crossover penalty.

\subsection{Focused Mode: Argmax-Shift as Empirical Confirmation}

In \texttt{focused} mode, KV wins $d_T$ at budgets $B = 4$ and
$B = 6$ with Mann-Whitney $r = -0.979$ and KS $D = 1.000$.  This is
the strongest counter-example in the experimental campaign: KQV's
better $K$ cache does not translate to better attention output when
one mis-quantised dominant key shifts the softmax argmax.

The mechanism is exactly as predicted in
Section~\ref{sec:kv_asymmetry}: concentrated attention amplifies
$K$ direction error exponentially via the softmax, and at high bit
budgets the additional transform step in WHT introduces a small but
decisive direction perturbation that KV's simpler scalar
reconstruction avoids.  The focused mode result is not a failure of
TurboQuant per se --- it is a certificate that the theoretical
analysis of Section~\ref{sec:kv_asymmetry} predicts the correct
failure mode in the correct regime.

\subsection{Fat-Tail $\nu$ Sweep: WHT as Distribution Normaliser}

We sweep $\nu \in \{10, 5, 3, 2, 1.5, 1.2, 1.05\}$ across all five
budgets, searching for a failure mode of KQV under extreme fat tails.
We do not find one.

The central finding is that KQV's $K$ reconstruction quality is
\emph{completely invariant} to $\nu$ at $B=4$:

\begin{center}
\begin{tabular}{rcccc}
\toprule
$\nu$ & \multicolumn{2}{c}{KQV} & \multicolumn{2}{c}{KV} \\
      & $\epsilon_K^{\mathrm{snr}}$ & $\epsilon_K^{\mathrm{dir}}$
      & $\epsilon_K^{\mathrm{snr}}$ & $\epsilon_K^{\mathrm{dir}}$ \\
\midrule
10   & 0.9345 & 0.0271 & 0.9372 & 0.0294 \\
 3   & 0.9342 & 0.0271 & 0.9666 & 0.0591 \\
 2   & 0.9345 & 0.0271 & 0.9802 & 0.0991 \\
1.5  & 0.9342 & 0.0271 & 0.9862 & 0.1419 \\
1.05 & 0.9341 & 0.0268 & 0.9902 & 0.2015 \\
\bottomrule
\end{tabular}
\end{center}

KQV's $\epsilon_K^{\mathrm{dir}}$ holds at $0.027$ to three
decimal places across the full range, while KV's grows $7\times$
from $0.029$ to $0.202$.  The mechanism is WHT acting as a
\emph{distribution normaliser}: whatever the marginal shape of the
input, the transform maps coordinates to approximately
Beta$(63.5,63.5)$ before the codebook sees them.  The heavier
the tail, the more work WHT does --- and the more protection it
provides.

At higher budgets ($B=5$) the effect is even cleaner: KQV's
$\epsilon_K^{\mathrm{dir}}$ actually \emph{improves} as $\nu$
decreases, from $0.024$ at $\nu=10$ to $0.010$ at $\nu=1.05$.
Near-basis-vector inputs after WHT produce a perfectly uniform
$\pm 1/\sqrt{d}$ pattern --- the best possible match to the Beta
codebook.  KV's direction error at $B=5$ stays near $0.044$
across all $\nu$, so the gap widens.

KV fails catastrophically at $B=2$ for $\nu \leq 1.5$:
\texttt{ptrelMSE\_K} $> 1$ (reconstruction error exceeds signal
energy), consistent with full codebook saturation on near-basis-vector
inputs.  KQV never triggers this warning at any $\nu$ or any budget.

The energy distance grows monotonically with tail weight (all
$p_{\mathrm{perm}} = 0.001$):

\begin{center}
\begin{tabular}{rcc}
\toprule
$\nu$ & $E_{6d}$ ($B=4$) & MW $r$ \\
\midrule
10   & 0.063 & $+0.307$ \\
 5   & 0.164 & $+0.650$ \\
 3   & 0.715 & $+1.000$ \\
 2   & 0.713 & $+1.000$ \\
1.5  & 0.902 & $+1.000$ \\
1.05 & 1.004 & $+1.000$ \\
\bottomrule
\end{tabular}
\end{center}

TurboQuant has no failure mode under fat tails.  The search was
honest and the parameter range extreme ($\nu = 1.05$ is near-Cauchy).
The negative result strengthens the theoretical claim: WHT protection
is unconditional for marginal concentration.

\section{The Low-Rank Pathology}

The most striking experimental result is not the QJL budget trade-off
but the behaviour of low-rank key distributions at high dimension.
Despite producing rotated coordinates that are indistinguishable from
the Beta distribution, low-rank keys cause catastrophic softmax
distortion — an order of magnitude worse than heavy-tailed keys.
This section analyses the mechanism and its practical consequences.

\begin{figure*}[htbp]
  \centering
  \includegraphics[width=\textwidth]{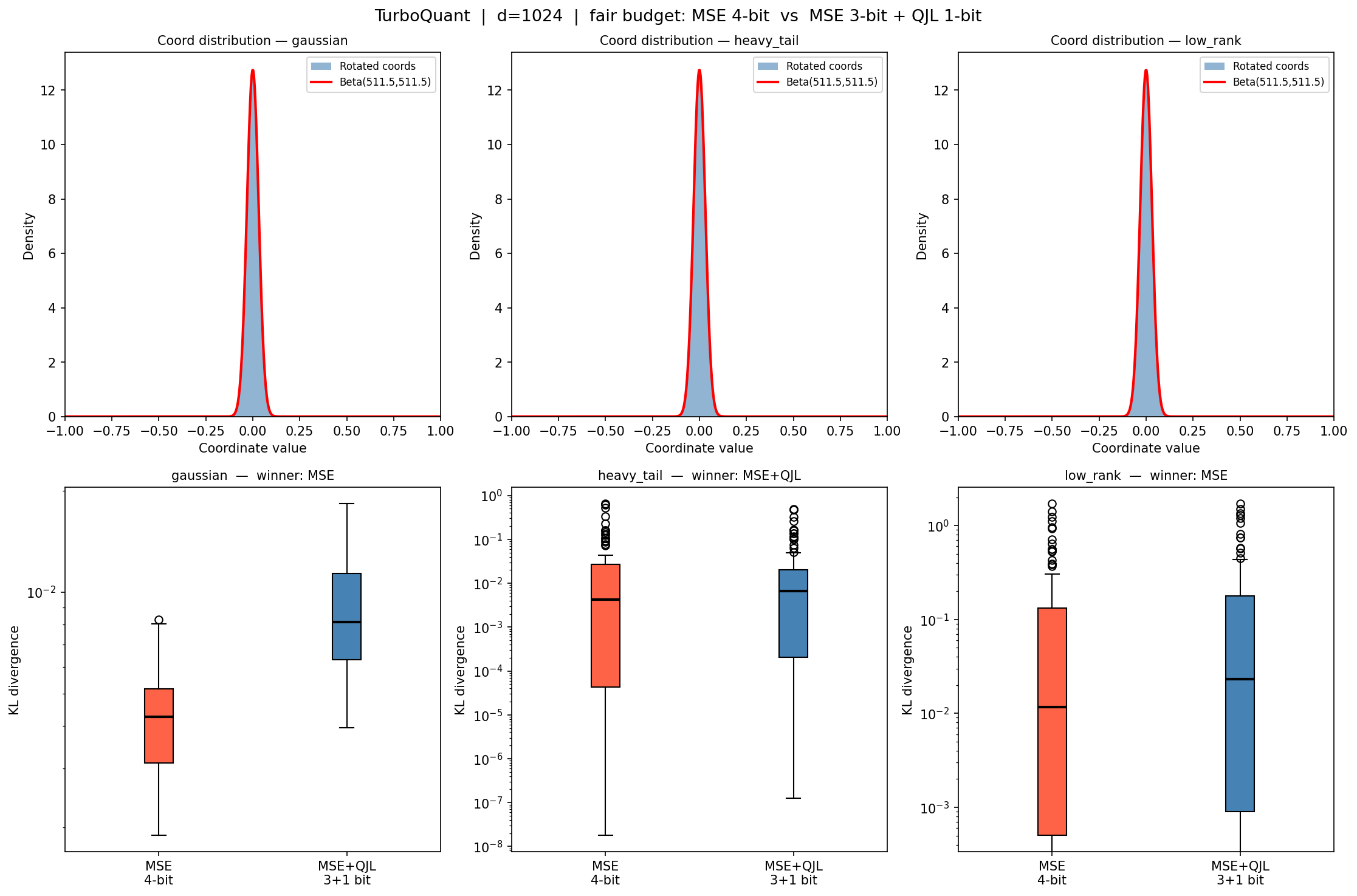}
  \caption{%
    $d=1024$, Beta$(511.5,511.5)$ — indistinguishable from
    $\mathcal{N}(0,1/d)$.
    \textit{Top}: all three distributions produce \emph{identical}
    histograms matching the Beta curve — the rotation completely hides
    the original structure.
    \textit{Bottom}: despite the perfect histogram match,
    \texttt{low\_rank} KL divergence is catastrophically large
    (${\sim}10^0$--$10^1$), exceeding \texttt{heavy\_tail} by one to
    two orders of magnitude. \texttt{heavy\_tail} is the only case where
    MSE+QJL wins under equal budget.
    Circles denote outliers beyond $1.5\times\mathrm{IQR}$
    (${\approx}2.7\sigma$ for a Gaussian).
  }
  \label{fig:d1024}
\end{figure*}

\subsection{The Histogram Paradox}
At $d=1024$, the histogram of rotated coordinates is identical for
gaussian, heavy\_tail, and low\_rank distributions: all match
Beta$(511.5, 511.5)$ perfectly (Figure~\ref{fig:d1024}, top row).
Yet the KL divergence for low\_rank is catastrophically large
($\sim 10^0$--$10^1$), far exceeding heavy\_tail ($\sim 10^{-1}$)
(Figure~\ref{fig:d1024}, bottom row).

This reveals that \textbf{Lemma~1 is a marginal guarantee}, not a joint
one. The rotation makes individual coordinates Beta-distributed, but the
low-rank structure survives in the correlations between coordinates.
The Lloyd-Max quantiser, applied coordinate-wise, assumes independence
and fails for joint-dependent distributions.

\subsection{Why Low-Rank is Worse than Heavy-Tail at High $d$}

\begin{observation}
At $d=1024$, the maximum KL divergence for low\_rank exceeds
that of heavy\_tail by one to two orders of magnitude.
\end{observation}

For \texttt{heavy\_tail}: normalization to the unit sphere neutralises
the large-norm outliers (norms stored separately); with 1024 dimensions
the CLT stabilises the inner product estimates.

For \texttt{low\_rank}: the data lies in a rank-$d/8$ subspace.
Queries and keys are strongly aligned within this subspace, producing
extremely peaked softmax distributions.
A small quantisation error on the dominant key is amplified
exponentially by the softmax, regardless of dimensionality.
No rotation or normalisation can fix this: the joint structure is
invisible in the marginal histogram but fully present in the inner
products.

\subsection{Practical Implication}
Transformer attention matrices are empirically low-rank — this is
exploited by LoRA~\cite{hu2022lora}, low-rank KV compression, and SVD-based methods.
\textbf{Low-rank is not an adversarial edge case; it is the realistic
  distribution of transformer KV caches.} TurboQuant's worst-case
behaviour is most severe precisely in these most common settings.

\section{Conclusions}

We surface several findings not prominent in the original TurboQuant
exposition:

\begin{enumerate}
  \item \textbf{QJL requires additive bits, not traded bits.}
        Under fair budget, MSE consistently outperforms MSE+QJL.
        QJL's benefit is bias elimination; its cost is variance
        injection proportional to $\|\mathbf{r}\|^2/d$.

  \item \textbf{Mean KL is an optimistic metric.}
        Worst-case KL divergence -- corresponding to rare but
        catastrophic attention corruption -- is the operationally
        relevant measure for KV cache quantisation reliability.

  \item \textbf{Lemma~1 guarantees marginals, not joints.}
        The Beta distribution is a necessary but not sufficient
        condition for good quantisation. Low-rank distributions
        produce matching histograms but catastrophic KL divergence.

  \item \textbf{Low-rank is more dangerous than heavy-tail at high $d$.}
        At $d=1024$, worst-case KL divergence for low-rank keys
        exceeds heavy-tailed keys by one to two orders of magnitude.
        Normalization and CLT stabilise heavy-tail inner products;
        no rotation or normalization can neutralise subspace
        concentration in the joint key distribution.

  \item \textbf{K--V asymmetry holds at the softmax level; KQV wins at $n=4$ on every measure.}
        The controlled comparison KQV vs QKQV holds $V$ identical between
        the two schemes, isolating the effect of QJL on $K$.
        WHT+scalar on $K$ imposes no inner product variance cost;
        QJL on $K$ inflates it by $(\pi/2)/\epsilon_B \geq \pi/2$, amplified
        nonlinearly by the softmax: KL divergence is consistently worse for
        QKQV at every budget and distribution.
        At the practically dominant budget $n=4$, KQV wins on every
        measure --- KL divergence, geometric $K$ reconstruction, and $d_6$
        --- across all ranks and tail weights tested.
        However, at $n \in \{2,3\}$, QKQV achieves better geometric $K$
        reconstruction (approaching the theoretical bound at $n=2$),
        a budget-dependent crossover invariant to rank and tail weight
        that the $\pi/2$ argument does not predict.

  \item \textbf{WHT provides unconditional protection against marginal
        concentration.}
        Across $\nu \in \{10,\ldots,1.05\}$ and all five budgets,
        KQV's $\epsilon_K^{\mathrm{dir}}$ holds at $0.027$ while
        KV's grows $7\times$.  The heavier the tail, the more WHT
        normalises the input to Beta before the codebook sees it.
        No failure mode was found in 35 experiments.

  \item \textbf{The rank failure mode is joint, not marginal.}
        Both KQV and KV maintain flat $K$ reconstruction quality
        across ranks $1$--$64$: the Beta marginal guarantee holds for
        both.  The failure emerges in the output $T$, where concentrated
        subspace structure peaks the softmax and amplifies small $K$
        errors exponentially.

\end{enumerate}

These findings suggest that robust KV cache quantisation requires
explicit handling of the low-rank structure of attention matrices ---
for example via SVD-based pre-processing, subspace-aware codebooks,
or selective full-precision preservation of dominant singular directions
--- and that worst-case KL divergence, not mean KL, should be the
primary evaluation metric for KV cache quantisation schemes.

\textbf{Open problem.}
The budget-dependent crossover (QKQV wins at $n \in \{2,3,5,7\}$,
KQV wins at $n \in \{4,6\}$) is confirmed invariant to both rank
and tail weight across two parameter sweeps.
The KL bridge of Section~\ref{sec:kl_bridge} provides a qualitative
mechanism: the scheme with lower K direction error produces lower KL
via Jensen's inequality, and the winner in KL determines the winner
in routing and output quality (Table~\ref{tab:kl_topk5}).
The crossover in KL tracks the crossover in geometric $K$ quality
exactly. What remains open is the \emph{quantitative} derivation:
under what condition does adding a 1-bit QJL residual corrector to
$(n-1)$-bit scalar quantisation produce lower K direction error than
$n$-bit scalar quantisation alone?
A complete theory would derive the crossover budget $n^*$ from the
rate-distortion curve of the Beta codebook and the JL residual
variance as a function of $n$ and $d$ --- a calculation we leave for
future work.

\textbf{Validation note.}
The current experiments are implemented in ROCm HIP on GPU with
simulation-side Python post-processing.  A planned next step is to
move all three quantisation schemes (KV, KQV, QKQV) fully onto GPU
and to increase the number of iterations per configuration to
eliminate sampling noise from the statistical estimates reported here.

\appendix

\section{Summary: Best Scheme by Regime and Budget}
\label{app:summary}

Table~\ref{tab:summary} collects the best-performing scheme across all tested
regimes and budgets, measured by the overall 6D error distance $d_6$.
The dominant pattern is a budget-driven alternation: QKQV wins at odd budgets
$n \in \{2,3,5,7\}$ and KQV wins at even budgets $n \in \{4,6\}$.
Two regimes produce exceptions where structural properties of the key
distribution override the alternation.

\begin{table}[htbp]
\centering
\small
\caption{%
  Best scheme per regime and budget (lowest $d_6$, overall 6D error distance).
  The alternating pattern QKQV / KQV at odd / even $n$ is the default.
  \textbf{Bold} marks deviations from that default.
  ``$\approx$'' denotes no statistically significant winner
  (permutation $p > 0.05$).
  All other cells: $p < 0.005$.}
\label{tab:summary}
\begin{tabular}{lcccccc}
\toprule
Regime & $n=2$ & $n=3$ & $n=4$ & $n=5$ & $n=6$ & $n=7$ \\
\midrule
Fat-tail ($\nu \geq 2$)
  & QKQV & QKQV & KQV & QKQV & KQV & QKQV \\
Fat-tail ($\nu \leq 1.5$, extreme)
  & QKQV & QKQV & KQV & KQV$^{\dagger}$ & KQV & QKQV \\
Low-rank ($r \leq 8$, degenerate)
  & QKQV & QKQV & \textbf{KV}$^{\ddagger}$ & QKQV & \textbf{KV}$^{\ddagger}$ & QKQV \\
Low-rank ($r \geq 16$, diffuse)
  & QKQV & QKQV & KQV$^{\S}$ & QKQV & \textbf{KV}$^{\ddagger}$ & QKQV \\
Random (isotropic Gaussian)
  & QKQV & $\approx$ & $\approx$ & QKQV & $\approx$ & QKQV \\
Focused (concentrated attention)
  & QKQV & QKQV & \textbf{KV}$^{\P}$ & QKQV & \textbf{KV}$^{\P}$ & QKQV \\
\bottomrule
\end{tabular}

\vspace{0.6em}
\noindent
$^{\dagger}$~At $\nu \leq 1.2$ (near-Cauchy) QKQV's QJL on $K$ compounds with
peaked attention to drive $\epsilon_T^{\mathrm{dir}}$ to $0.23$ vs.\ $0.08$
for KQV; KQV wins at $n=5$ as well as $n=4,6$.

\smallskip\noindent
$^{\ddagger}$~Full ranking: KV $>$ KQV $>$ QKQV.
When $K$ lies near a low-dimensional subspace, WHT's $K$-direction benefit
disappears (both schemes achieve the same $\epsilon_K^{\mathrm{dir}}$), but
KQV still pays the $(n-1)$-bit penalty on $V$.
Plain scalar quantisation wins both components.

\smallskip\noindent
$^{\S}$~KQV marginally beats KV and QKQV (MW $r \approx 0.15$, $p < 0.05$);
effect size small and shrinks further at $r=64$.

\smallskip\noindent
$^{\P}$~Full ranking: KV $>$ KQV $>$ QKQV.
Concentrated attention amplifies $K$ direction error exponentially
via the softmax argmax; KV's simpler reconstruction avoids WHT's
additional direction perturbation at high budgets.
\end{table}

\paragraph{Reading the table.}
Every row where the cell reads QKQV or KQV (no bold) follows the
alternating-budget rule established in Section~\ref{sec:factorial}.
The bold KV cells mark the two operating conditions under which WHT's
protection is absent (low-rank degenerate keys, focused queries): in
both cases the ranking inverts to KV $>$ KQV $>$ QKQV at even budgets,
while reverting to QKQV $>$ KQV $>$ KV at odd budgets.
The random (isotropic) regime is the boundary case: WHT neither helps
nor hurts $K$ reconstruction, and the QKQV/KQV differences at even $n$
are not statistically distinguishable from zero.

\end{document}